\documentclass[11pt]{article}

\usepackage[final]{acl}

\usepackage{times}
\usepackage{latexsym}

\usepackage[T1]{fontenc}

\usepackage[utf8]{inputenc}

\usepackage{microtype}

\usepackage{inconsolata}

\usepackage{graphicx}

\usepackage{tabularx} 
\usepackage{booktabs}
\usepackage{xcolor}
\usepackage{pifont}    
\usepackage{caption}
\usepackage{geometry}
\usepackage{graphicx}
\usepackage{multirow}
\usepackage[table]{xcolor}
\usepackage{float}  
\usepackage{minted}
\usepackage{float}
\usepackage{placeins}
\usepackage{float}

\usepackage[utf8]{inputenc} 
\usepackage[T1]{fontenc}    
\usepackage{hyperref}
\usepackage{url}
\usepackage{booktabs}       
\usepackage{amsfonts}       
\usepackage{nicefrac}       
\usepackage{microtype}      
\usepackage{xcolor}         
\usepackage{amsmath}  
\usepackage[ruled,vlined]{algorithm2e} 
\usepackage{algpseudocode} 
\usepackage{graphicx} 
\usepackage{wrapfig} 
\usepackage{booktabs}    
\usepackage{multirow}    
\usepackage{array}       
\usepackage[table]{xcolor} 
\usepackage{caption}    
\usepackage{subcaption}

\usepackage{array}
\usepackage{tabularx}
\usepackage{booktabs}
\usepackage{setspace} 
\usepackage{mathtools}
\usepackage[labelformat=empty]{subcaption}
\usepackage{tcolorbox}
\usepackage{cuted}

\usepackage{mathtools}

\definecolor{Gray}{gray}{0.9}

\usepackage{xcolor}
\usepackage{xfp}
\usepackage{xcolor}
\usepackage{colortbl}
\definecolor{rpolight}{RGB}{245,248,255}

\usepackage{multicol}
\usepackage{float} 
\usepackage{graphicx}

\newcommand{\valdiff}[2]{%
#1\hspace{0.3em}%
{\scriptsize
\edef\diffval{\fpeval{round(#1-(#2),1)}}%
\ifdim\diffval pt>0pt
  \textcolor{green!60!black}{+\diffval}%
\else
  \textcolor{red}{\diffval}%
\fi
}%
}

\title{RPO:Reinforcement Fine-Tuning with Partial Reasoning Optimization}

\author{
Hongzhu Yi, Xinming Wang, Zhenghao zhang, Tianyu Zong, Yuanxiang Wang, Jun Xie, Tao Yu, Haopeng Jin \\ \textbf{Kaixin Xu, Feng Chen, Jiahuan Chen, Yujia Yang, Zhenyu Guan, Bingkang Shi, Jungang Xu*} \\
\texttt{\{yihongzhu23, zhangzhenghao25, guanzhenyu24\}@mails.ucas.ac.cn}\\
\texttt{\{zongtianyu20,yangyujia24, wangyuanxiang19\}@mails.ucas.ac.cn} \\
\texttt{\{wangxinming2024,yutao2025\}@ia.ac.cn} \\
\texttt{orangenius@sjtu.edu.cn} \\
\texttt{\{xiejun,chenfeng13\}@lenovo.com} \\
\texttt{sunbeam.King@bupt.edu.cn} \\
\texttt{xujg@ucas.ac.cn}
}

\begin{document}
\maketitle
\begin{abstract}

Within the domain of large language models, reinforcement fine-tuning algorithms necessitate the generation of a complete reasoning trajectory beginning from the input query, which incurs significant computational overhead during the rollout phase of training. To address this issue, we analyze the impact of different segments of the reasoning path on the correctness of the final result and, based on these insights, propose Reinforcement Fine-Tuning with Partial Reasoning Optimization (\textbf{RPO}), a plug-and-play reinforcement fine-tuning algorithm. Unlike traditional reinforcement fine-tuning algorithms that generate full reasoning paths, RPO trains the model by generating suffixes of the reasoning path using experience cache.
During the rollout phase of training, RPO reduces token generation in this phase by approximately 95\%, greatly lowering the theoretical time overhead. Compared with full-path reinforcement fine-tuning algorithms, RPO reduces the training time of the 1.5B model by 90\% and the 7B model by 72\%. At the same time, it can be integrated with typical algorithms such as GRPO and DAPO, enabling them to achieve training acceleration while maintaining performance comparable to the original algorithms. Our code is open-sourced at \url{https://github.com/yhz5613813/RPO}.

\end{abstract}

\section{Introduction}
\label{sec:intro}
In recent years, large language models (LLMs) \citep{openai2024gpt4technicalreport,touvron2023llamaopenefficientfoundation,zeng2023glm130bopenbilingualpretrained} have achieved remarkable breakthroughs in reasoning and generalization capabilities~\citep{wang2025hitchhiker}, particularly after the introduction of reinforcement learning during the post-training stage \citep{ouyang2022traininglanguagemodelsfollow}. Pioneering works such as OpenAI's O1 \citep{openai2024openaio1card} and DeepSeek-R1 \citep{deepseekai2025deepseekr1incentivizingreasoningcapability} have demonstrated impressive reasoning-time efficiency, primarily due to the synergistic combination of reinforcement learning and chain-of-thought (CoT) reasoning \citep{wei2023chainofthoughtpromptingelicitsreasoning}. This paradigm shift highlights the transformative potential of Reinforcement Learning-based post-training in pushing the boundaries of LLM performance.

Despite its promising prospects, applying reinforcement learning in post-training remains immature. In terms of time overhead, Reinforcement Learning fine-tuning typically generates a large number of samples during the sampling stage. However, parameter updates cannot proceed until all samples are completed, resulting in severe underutilization of computational resources. Although some asynchronous Reinforcement Learning approaches exist, the off-policy nature of the optimization process requires more training iterations. Furthermore, during Reinforcement Learning fine-tuning of language models, rewards are usually computed only after generating the final token based on task-specific criteria. This paradigm, known as Reinforcement Learning with Verifiable Rewards \citep{lambert2025tulu3pushingfrontiers, Wang_2025}, lacks intermediate feedback and produces sparse rewards. Such sparsity hinders the model’s ability to learn optimal policies and contributes to training instability \citep{lightman2023letsverifystepstep}.

We observe that a significant underlying issue stems from the policy model's need to identify a reasoning trajectory from the beginning of the problem to the correct answer. This approach—comparing entire reasoning paths using policy gradients—leads to excessive randomness during the sampling phase. Although it expands the search space, it often fails to find suitable reasoning paths, resulting in inefficient sampling and high variance.

\begin{figure}
    \centering
    \includegraphics[width=1\linewidth]{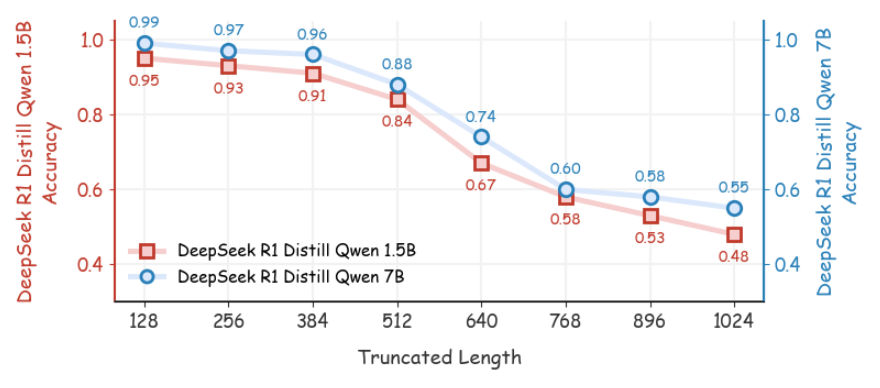}
    \includegraphics[width=1\linewidth]{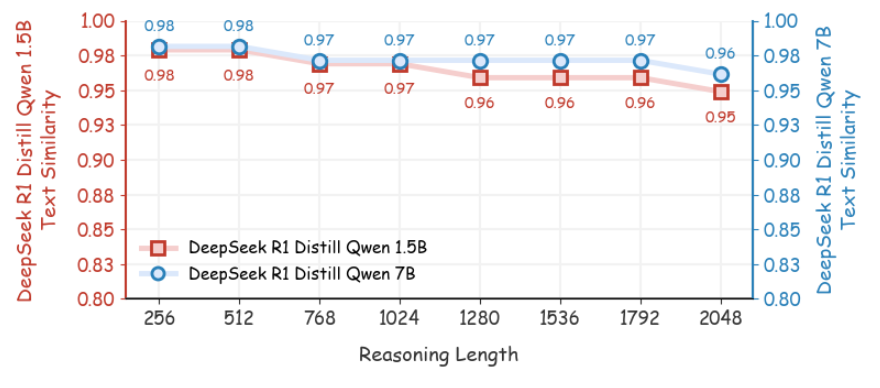}
    \caption{The top figure shows the \textbf{DeepSeekR1-Qwen-Distill-7b} and \textbf{DeepSeekR1-Qwen-Distill-1.5b} models. For each question, an initial answer is generated and then truncated; from the truncation point, 256 answers are subsequently generated, and the relationship between truncation length and the overall average accuracy is analyzed. The bottom figure shows 256 answers generated for each training question. Answers exceeding 2048 tokens are selected, and \textbf{BERT} is used to measure the similarity between equal-length prefix segments. The similarity metric is defined as: $
\text{sim} = \frac{2}{n(n-1)} \sum_{i=1}^{n-1} \sum_{j=i+1}^{n} \frac{\text{BERT}(s_i) \cdot \text{BERT}(s_j)^\top}{\|\text{BERT}(s_i)\| \, \|\text{BERT}(s_j)\|}$.}
    \label{fig:pretoken_combined}
\end{figure}

An alternative perspective arises: since exploring a complete reasoning path from the beginning of the problem introduces various drawbacks, why not train the policy model to complete a reasoning path based on \textbf{partially correct reasoning process hint} instead?
We found that this is feasible. Through our experiments, enabling the model to complete correct reasoning paths can still effectively teach it to generate whole reasoning trajectories from the initial problem statement. Based on this insight, we propose the \emph{\textbf{R}eplay-based \textbf{P}olicy \textbf{O}ptimization}(RPO). Our method is grounded in a reasonable assumption: the early tokens of a reasoning path that leads to the correct answer are more likely to guide the model toward the correct reasoning trajectory. Furthermore, we investigate the relationship between the length of the truncated trailing tokens and the model's generation accuracy. As shown in Figure~\ref{fig:pretoken_combined}, the initial tokens of correct answers play a crucial role in guiding the model toward correct solutions, and longer prefix lengths are positively correlated with higher generation accuracy.

Specifically, we construct a cache pool for reinforcement fine-tuning to store previously generated reasoning paths and continuously update it during training. After completing the sampling generation stage for each question, we add the reasoning path that leads to the correct answer into the cache. When the same question is encountered again, we retrieve the first $n$ tokens of the corresponding reasoning path from the cache, prepend them to the prompt, and then perform sampling generation. At the same time, we design a reward function that adaptively adjusts based on response length to improve the accuracy of gradient estimation in the experience caching scenario.

Experimental results show that this method is plug-and-play, concise and effective, enhances training stability during the reinforcement learning stage, significantly reduces the policy model’s sampling time cost, and achieves performance improvements.

We propose RPO, a novel framework for reinforcement fine-tuning of LLMs, introducing an experience replay mechanism in the sampling stage. Key advantages are: \textbf{plug-and-play}: allowing easy integration into other Reinforcement Learning fine-tuning methods; \textbf{reduced resource consumption}: achieving up to 92.6\% faster training; \textbf{strong stability}: mitigating common Reinforcement Learning instability in reasoning models.We evaluated RPO on Deepseek-R1-Distill-Qwen 1.5B and 7B models across six datasets. The results show that training time was reduced by approximately 90\%, and compared to full-path exploration in GRPO and DAPO, performance improved by about 2\%.

\section{Related Work}
\label{gen_inst}
\paragraph{Reinforcement Fine-Tuning.}
Reinforcement Fine-Tuning (RFT) guides the model fine-tuning process through the reward mechanisms of reinforcement learning, greatly enhancing generalization and accuracy. Kimi v1.5 \citep{team2025kimi} and ReFT \citep{luong2024reft} employ traditional Proximal Policy Optimization (PPO) \citep{schulman2017proximal} for RFT and have demonstrated excellent performance. DeepSeek-R1 \citep{deepseekai2025deepseekr1incentivizingreasoningcapability} adopts GRPO and uses verifiable reward strategies to compute policy gradients directly. DAPO \citep{yu2025dapoopensourcellmreinforcement} further optimizes GRPO to improve training stability.

\paragraph{Experience Replay.}
Recently, a small portion of work has also explored experience replay. For example, TreePO\cite{li2025treepobridginggappolicy} organizes sequences into tree structures, accelerating the rollout phase and achieving some performance improvement. SegmentPO \cite{guo2025segmentpolicyoptimizationeffective} leverages subtree information to provide process supervision for each node. BREAD \cite{zhang2025breadbranchedrolloutsexpert} combines the SFT and RFT stages and attempts to incorporate prefixes of correct answers when reasoning fails. However, these approaches focus on tree construction, resulting in relatively complex algorithms, and the acceleration in reasoning is not particularly significant. RPO focuses on improving the accuracy of gradient estimation through a minimal revision of the reward function, and combines this with experience caching to achieve simultaneous improvements in training speed and performance.

\begin{figure*}[ht]
	\centering
 	\includegraphics[width=\textwidth]{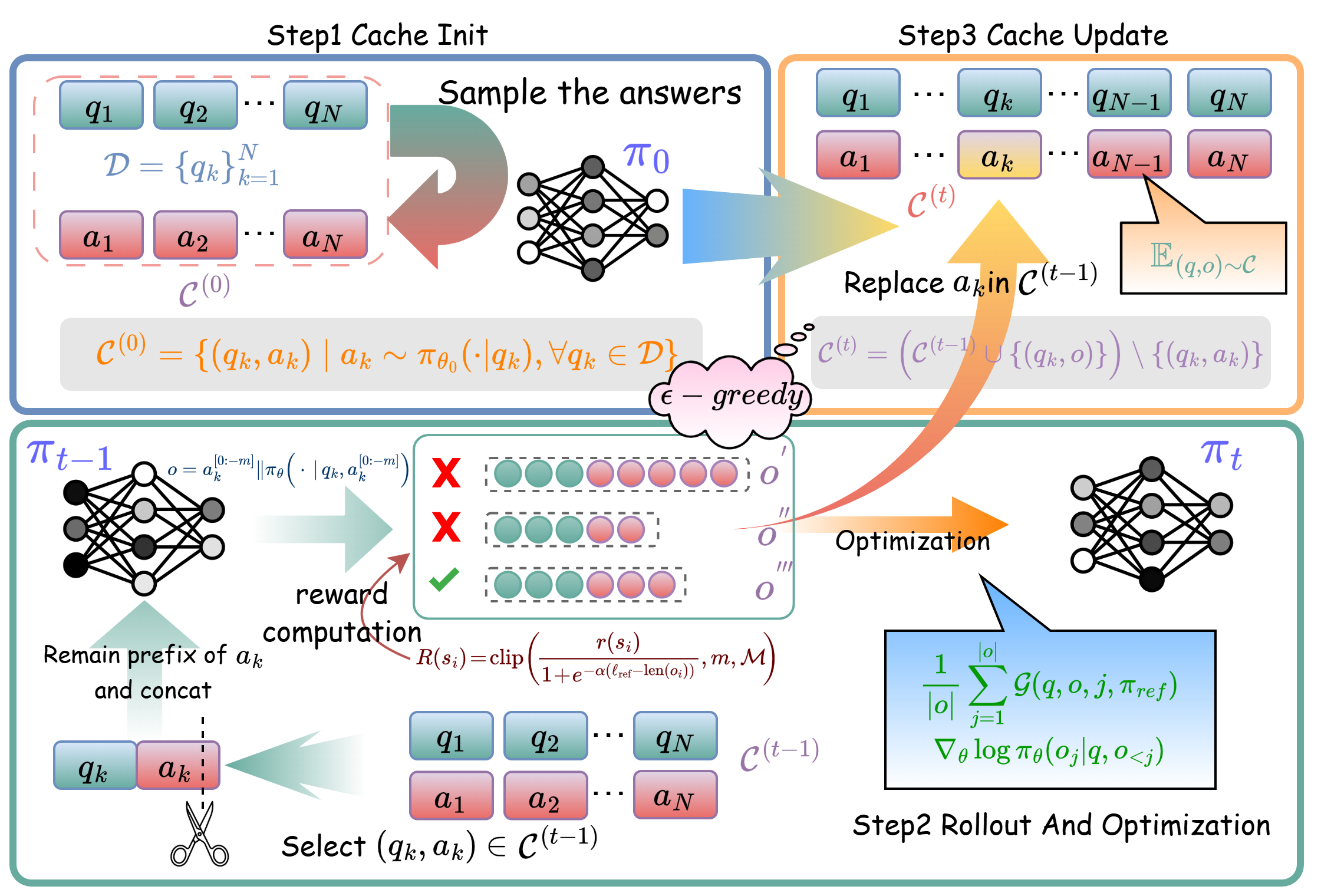}
	\caption{Overview of the RPO framework. The entire training process is described as follows: Cached answer fragments are used by the model to generate new responses; either the best or a random response is selected based on the reward system for optimization; and the cache is continuously updated to improve training efficiency and stability.}
	\label{fig:overview}
    \vspace{-15pt}
\end{figure*}

\section{Method}
\label{method}
The overall framework of our method is illustrated in Figure ~\ref{fig:overview} and consists of three main components. The first component involves cache initialization. In the second component, subsequent responses are rolled out based on the initialized cache and subsequently optimized. The third component updates the cache pool using an $\epsilon$-greedy strategy.

\subsection{{Cache Pool Initialization}}

First, we denote the dataset of samples as \(\mathcal{D} = \{q_k\}_{k=1}^N\), where \(q_k\) represents the \(k\)-th question in the dataset. We denote the initial model parameters as \(\theta_0\), and we represent the model’s answering policy by \(\pi_{\theta_0}\). Before training begins, we initialize the cache pool as \(\mathcal{C}^{(0)}\) as follows:
\begin{equation}
\mathcal{C}^{(0)} = \left\{ \left( q_k, a_k \right) \mid a_k \sim \pi_{\theta_0}(\cdot|q_k), \forall q_k \in \mathcal{D} \right\}.
\end{equation}

This stage uses the initial model policy to sample the dataset $\mathcal{D}$. To retrieve the response $a_k$ corresponding to question $q_k$ from the cache pool, we define the retrieval operation as:
\begin{equation}
a_k \coloneqq \{ a \mid (q_k, a) \in \mathcal{C} \}.
\end{equation}
Here, $a_k$ denotes the answer associated with question $q_k$ in the cache pool $\mathcal{C}$.

\subsection{Rollout And Optimization}
\label{truncated answer sampling optimization}

\textbf{Rollout.} At each rollout stage, we use the RPO strategy to retrieve the historical response \(a_k\) for each question \(q_k\) from the cache pool \(\mathcal{C}\). We then remove the last \(m\) tokens and concatenate the remaining prefix with \(q_k\) to construct the input instruction, thereby generating a new response \(o\). We express this process as:

\begin{equation}
o = a_k^{[0:-m]} \Vert \pi_\theta\big(\cdot \,|\, q_k, a_k^{[0:-m]}\big),
\end{equation}
where $
    a_k \! \coloneqq \! \{ a \! \mid \! (q_k, a) \! \in \! \mathcal{C} \} ,
    m \! \sim \! \mathcal{U}\{0,\! 1, ..., \! L\}.
$,$L$ is the maximum truncation length, $\mathcal{U}\{0,1,..., L\}$ samples a truncation point uniformly from $[0, L]$, $a_k^{[0:-m]}$ truncates the last $m$ tokens of $a_k$, and $\pi_\theta(\cdot|q_k, a_k^{[0:-m]})$ generates a new continuation based on the question and prefix. In this paper, $L$ is either fixed or set dynamically based on the shortest response in a sampling group $G$, denoted as $\ell$, where:
\begin{equation}
    \ell = \min\{\mathrm{len}(o_1), \mathrm{len}(o_2), \dots, \mathrm{len}(o_G)\}.
\end{equation}


\noindent{\textbf{Replay-based Policy Optimization.}} After completing the sampling generation, RPO adopts \textbf{Group Relative} estimation of advantage. For a given question-answer pair $(q_k, a_k)$, the behavioral policy $\pi_{\theta_{t-1}}$ samples a group of $G$ individual responses $\{o_i\}_{i=1}^G$ from the model. Then, by normalizing the group rewards $\{R_i\}_{i=1}^G$, the advantage of each response is computed as:
\begin{align}
&\mathcal{J}_{\mathrm{RPO}}(\theta_t)
=
\mathbb{E}_{(q,a)\sim \mathcal{C}^{(t-1)},\,o_{1:G}\sim \pi_{\theta_t}}\nonumber\\
&\left[\! 
\frac{1}{G}\sum_{i=1}^G \!
\frac{1}{|o_i|} \!
\sum_{j=1}^{|o_i|} \!
\ell_{i,j}(\theta_t) \!
\right] \!
- \!
\beta\, \! D_{\mathrm{KL}}\!\left(\pi_{\theta_t}\, \! \|\, \!\pi_{\mathrm{ref}}\right),
\end{align}

\noindent
where the token-level RPO loss is defined as:
\begin{equation}
\begin{split}
    \ell_{i,j}(\theta_t) = \min \Big( & r_{i,j}(\theta_t)\hat{A}_{i,j}, \\
    & \mathrm{clip}\big(r_{i,j}(\theta_t), 1\pm\epsilon\big)\hat{A}_{i,j} \Big).
\end{split}
\end{equation}

\noindent
The probability ratio and the normalized advantage are given by
\begin{equation}
\!\! r_{i,j}(\! \theta_t \!) \!
= \!
\frac{\pi_{\theta_t}(o_{i,j} \! \mid \! q,o_{i,<j})}
{\pi_{\theta_{t-1}}(o_{i,j} \! \mid \! q,o_{i,<j})},
\hat{A}_{i,j} \!
= \!
\frac{R_i \! - \! \mu_R}{\sigma_R},
\end{equation}
where $\mu_R$ and $\sigma_R$ denote the mean and standard deviation of
$\{R_i\}_{i=1}^G$, respectively.

\vspace{1em}

Since RPO is policy-agnostic, we propose a unified forward reinforcement learning paradigm based on experience replay. We can then write the policy gradient function of RPO in a more general form as:
\begin{align}
&\nabla_{\theta} \mathcal{J}_{\mathrm{RPO}}(\theta) 
= \underbrace{\mathbb{E}_{(q,o) \sim \mathcal{C}}}_{\text{Data Source}} \nonumber \\
& \Bigg( \!\!
\frac{1}{|o|} \!\! \sum_{j=1}^{|o|} \!
\underbrace{\mathcal{G}(q,o,j,\pi_{ref})}_{\text{Gradient Coefficient}} \!
\nabla_{\theta} \! \log \pi_{\theta}(o_j | q, o_{<j}) \!\!
\Bigg)
\label{eq:normal J}
\end{align}

Equation~\ref{eq:normal J} is derived from the standard policy gradient formulation. The above equations indicate that only the sampling stage is affected by RPO, while the policy gradient function remains unaltered. As a result, RPO exhibits a plug-and-play nature and can be easily integrated into other reinforcement fine-tuning algorithms.

Compared with traditional full-path reasoning optimization, RPO introduces previously sampled historical response trajectories as constraints during subsequent sampling. In this way, the policy space \( \pi_{\theta} \) explored during training is restricted. This constraint regularizes the gradient descent space during learning, which can be expressed as:

\begin{equation} \!\! Var(\|\nabla_{\theta}\mathcal{J}_{\mathrm{RPO}}\|_2)\leq Var(\|\nabla_{\theta} \mathcal{J}_{\mathrm{ALL}}\|_2). \end{equation}
where,\(\nabla_{\theta} \mathcal{J}_{\mathrm{ALL}}\) represents the gradient of the full-path reasoning optimization. Theoretically, our method enables a more stable training process. Detailed mathematical proofs are provided in Appendix~\ref{sec:proof}.

\subsection{Cache Update}
\label{cache pool update}

After each gradient update, we adopt the \(\varepsilon\)-greedy algorithm to update the experience cache by selecting the highest-reward response from the current inference results. Specifically, when the random variable \( u \sim \mathcal{U}(0,1) \) satisfies \( u \leq \varepsilon \), we select the response with the highest group reward; otherwise, we randomly select a suboptimal response. We formalize the update process as:
\begin{equation}
\mathcal{C}^{(t)} = 
\bigl( \mathcal{C}^{(t-1)} \cup \{ (q_k, o) \} \bigr) 
\setminus \{ (q_k, a_k) \},
\end{equation}
where \( o = o_{\text{argmax}\{R_i\}_{i=1}^G} \) if \( u \le \varepsilon \), and \( o = o_{g'} \) otherwise. Here, \( o_{\text{argmax}\{R_i\}_{i=1}^G} \) denotes the highest-reward response in the group, and \( o_{g'} \) is another randomly selected candidate response.

\subsection{Length-Aware Reward Shaping}
However, since the same response prefix is shared during the sampling phase, the diversity of responses within the group is reduced compared to the full-path reasoning optimization algorithm. This lower diversity results in more similar reward signals, thereby diminishing the effectiveness of policy gradient estimation. To ensure meaningful gradients, reasonable reward differences should be maintained within the group, even when all responses are correct.

A \textbf{Length-Aware Reward Shaping} method is proposed to address this issue. This method is based on the assumption: \textit{For the same question, a reasoning path that reaches the correct answer more concisely should be rewarded with a higher value.} Specifically, for each response \( o_i \) in the group, its length-aware reward \( R(s_i) \) is computed as:
\begin{equation}
\!\!\! R(s_i) \! = \! \text{clip} \! \left( \! \frac{r(s_i)}{1 \! + \! e^{-\alpha (\ell_{\text{ref}} - \text{len}(o_i))}}, m, \mathcal{M} \! \right ),
\end{equation}
where, \( r(s_i) \) is the original reward, \( \ell_{\text{ref}} \) is the average length of group \( G \), defined as \( \ell_{\text{ref}} = \frac{1}{|G|} \sum_{i=1}^{|G|} \text{len}(o_i) \). The parameter \( \alpha > 0 \) controls the sensitivity of the reward to length differences. \( m \) and \( \mathcal{M} \) are the lower and upper bounds for reward clipping to avoid extremely large or small values. \( \text{clip}(\cdot, m, \mathcal{M}) \) denotes restricting a value within the interval \( [m, \mathcal{M}] \).

We then iterate the above steps in Sections~\ref{truncated answer sampling optimization} and ~\ref{cache pool update} until a predefined stopping step \( T \) is reached.

Through mathematical derivation, we demonstrate that length-aware rewards are better suited for the RPO algorithm; the two can complement each other, and when the guiding path is within a certain threshold, they can enable the model to achieve greater performance gains. In contrast, traditional GRPO or DAPO algorithms, which use the full reasoning path, lack an initial fixed guiding path. This results in high variance for length-aware rewards, making it difficult to accurately estimate the true effective policy gradient, and thus they are not suitable for using length-aware rewards. Detailed proofs are provided in Appendix~\ref{proof:2}.

\section{Experiments}

\subsection{Experimental Setup}
\label{Experimental Setup}
\paragraph{Base Models.}
To demonstrate the effectiveness and generality of RPO, we evaluate it on two open-source inference models with 1.5B and 7B parameters, namely \textbf{Deepseek-r1-qwen-distill-1.5b} and \textbf{Deepseek-r1-qwen-distill-7b}~\citep{deepseekai2025deepseekr1incentivizingreasoningcapability,qwen}. Notably, we skip the supervised fine-tuning (SFT) phase, which is usually a prerequisite for reinforcement learning to enhance performance~\citep{chu2025sftmemorizesrlgeneralizes}, as the selected models have already undergone this stage~\citep{deepseekai2025deepseekr1incentivizingreasoningcapability}.

\paragraph{Datasets.}
We evaluate the models on six standard reasoning evaluation datasets: \texttt{aime25}\citep{AIME25}, \texttt{aime24}\citep{AIME24}, \texttt{math500}\citep{hendrycks2021measuringmathematicalproblemsolving}, \texttt{amc23}\citep{AMC23}, \texttt{minerva}\citep{lewkowycz2022solving} and \texttt{olympicbench}\citep{he2024olympiadbenchchallengingbenchmarkpromoting}. To ensure fairness, all evaluations use the \texttt{lighteval}\citep{lighteval} toolkit.

\paragraph{Implementation Details.}
During training, we use 7k samples from the \texttt{open-rs} dataset~\citep{dang2025reinforcementlearningreasoningsmall} with a global batch size of 576 for 4 epochs. Experiments are run on a single H20 machine with 8×H20 96G GPUs. We generate 6 samples per prompt, set the temperature to 0.7, and fix the maximum generation length at 4096. For the length-aware reward, we use $m=0.5$, $M=1$, and $\alpha=0.01$. All models are fully fine-tuned. Due to time constraints, only zero-shot performance is averaged over three runs; all other ablation experiments are run once.

\subsection{Zero-shot performance}
We set the maximum truncation length $L$ for each group to half of the minimum response length $\ell$. 
Then, we train the DeepSeek-R1-Qwen-1.5B and DeepSeek-R1-Qwen-7B models for four epochs 
using the original GRPO algorithm, the DAPO algorithm, as well as their RPO variants, 
with a batch size of 576 and a maximum generation length of 4096 tokens. 
To ensure that the experimental results are not caused by randomness, 
we repeat the training three times for each experiment, then compare their mean accuracy on the designated evaluation datasets.

\begin{table*}[ht]
\centering \normalsize \setlength{\tabcolsep}{5pt}
\renewcommand{\arraystretch}{0.85}
\caption{Performance of the RPO algorithm on test datasets. w/ R means length-aware reward is used, w/o R means length-aware reward is not used. +RPO shows the effect of applying the RPO algorithm on top of each method.}

\resizebox{\linewidth}{!}{
\begin{tabular}{lccccccr}
\toprule
\textbf{Model} & \textbf{AIME25} & \textbf{AIME24} & \textbf{MATH500} & \textbf{AMC23} & \textbf{Minerva} & \textbf{OlyB} & \textbf{Avg} \\
\midrule

\multicolumn{8}{l}{\textit{\color{gray}{1.5B Models}}} \\
\midrule

DeepSeek-R1-Qwen-1.5B &
16.7 & 28.8 & 82.2 & 62.9 & 26.5 & 43.3 & 43.4 \\

\quad + GRPO (w/o R) &
24.4 & 31.1 & 85.7 & 72.5 & 29.8 & 51.3 & 49.1 \\

\rowcolor{rpolight}
\quad\quad +RPO &
\valdiff{24.4}{24.4} &
\valdiff{25.6}{31.1} &
\valdiff{84.3}{85.7} &
\valdiff{69.2}{72.5} &
\valdiff{29.5}{29.8} &
\valdiff{51.7}{51.3} &
\valdiff{47.5}{49.1} \\

\quad + GRPO (w/ R) &
22.2 & 32.2 & 83.8 & 70.8 & 27.5 & 50.5 & 47.8 \\

\rowcolor{rpolight}
\quad\quad +RPO &
\valdiff{24.4}{22.2} &
\valdiff{35.6}{32.2} &
\valdiff{85.3}{83.8} &
\valdiff{83.3}{70.8} &
\valdiff{29.8}{27.5} &
\valdiff{51.8}{50.5} &
\valdiff{51.7}{47.8} \\

\quad + DAPO (w/o R) &
30.0 & 24.4 & 86.2 & 84.2 & 29.7 & 52.7 & 51.2 \\

\rowcolor{rpolight}
\quad\quad +RPO &
\valdiff{28.9}{30.0} &
\valdiff{24.4}{24.4} &
\valdiff{86.0}{86.2} &
\valdiff{84.5}{84.2} &
\valdiff{29.3}{29.7} &
\valdiff{52.1}{52.7} &
\valdiff{50.9}{51.2} \\

\quad + DAPO (w/ R) &
26.7 & 30.0 & 85.0 & 84.1 & 29.7 & 51.1 & 50.2 \\

\rowcolor{rpolight}
\quad\quad +RPO &
\valdiff{32.2}{26.7} &
\valdiff{30.0}{30.0} &
\valdiff{86.2}{85.0} &
\valdiff{86.1}{84.1} &
\valdiff{29.1}{29.7} &
\valdiff{52.3}{51.1} &
\valdiff{52.7}{50.2} \\

\midrule
\multicolumn{8}{l}{\textit{\color{gray}{7B Models}}} \\
\midrule

DeepSeek-R1-Qwen-7B &
43.3 & 55.5 & 92.8 & 90.0 & 44.5 & 67.4 & 65.6 \\

\quad + GRPO (w/o R) &
43.3 & 53.3 & 95.0 & 90.0 & 44.5 & 67.2 & 65.6 \\

\rowcolor{rpolight}
\quad\quad +RPO &
\valdiff{43.3}{43.3} &
\valdiff{46.6}{53.3} &
\valdiff{92.5}{95.0} &
\valdiff{89.2}{90.0} &
\valdiff{42.3}{44.5} &
\valdiff{67.7}{67.2} &
\valdiff{63.6}{65.6} \\

\quad + GRPO (w/ R) &
40.0 & 48.9 & 95.0 & 88.3 & 43.5 & 66.0 & 63.6 \\

\rowcolor{rpolight}
\quad\quad +RPO &
\valdiff{50.0}{40.0} &
\valdiff{61.1}{48.9} &
\valdiff{94.2}{95.0} &
\valdiff{90.8}{88.3} &
\valdiff{43.7}{43.5} &
\valdiff{67.3}{66.0} &
\valdiff{67.8}{63.6} \\

\quad + DAPO (w/o R) &
43.3 & 53.3 & 94.6 & 90.2 & 45.1 & 67.7 & 65.7 \\

\rowcolor{rpolight}
\quad\quad +RPO &
\valdiff{46.7}{43.3} &
\valdiff{52.2}{53.3} &
\valdiff{94.2}{94.6} &
\valdiff{91.2}{90.2} &
\valdiff{42.7}{45.1} &
\valdiff{64.9}{67.7} &
\valdiff{65.3}{65.7} \\

\quad + DAPO (w/ R) &
42.2 & 56.7 & 93.2 & 91.8 & 44.6 & 64.5 & 65.5 \\

\rowcolor{rpolight}
\quad\quad +RPO &
\valdiff{46.7}{42.2} &
\valdiff{54.5}{56.7} &
\valdiff{94.8}{93.2} &
\valdiff{95.2}{91.8} &
\valdiff{43.1}{44.6} &
\valdiff{64.5}{64.5} &
\valdiff{66.5}{65.5} \\

\bottomrule
\end{tabular}
}
\vspace{-15pt}
\label{tab:overall_performance}
\end{table*}

As shown in Table~\ref{tab:overall_performance}, as mentioned in the Method section, length-aware rewards complement the RPO algorithm. Incorporating group-wise length-aware rewards enables RPO to achieve higher accuracy on test benchmarks than GRPO and DAPO for both the 1.5B and 7B model sizes. Without group-wise length-aware rewards, RPO may experience some performance degradation; therefore, when using RPO for accelerated training, it is recommended to include group-wise length-aware rewards to enhance performance.

\subsection{Training Time Overhead}

To investigate the training time overhead of GRPO and RPO, each experiment is conducted on a machine with 8 H20 GPUs, using only a single GPU for sampling during the training phase. It should be noted that RPO introduces additional inference overhead during the cache initialization phase, where parallel inference is performed across all GPUs using the \texttt{vllm} framework. When the dataset size is 7k and the parallel batch size is 256, this phase takes approximately 20 minutes.
Our experiments reveal that the primary factors affecting the relative training speed between RPO and GRPO are the number of group samples $G$ and the maximum truncation length $L$, while the impact of batch size is relatively minor.

\begin{table}[t!]
    \centering
    \caption{The average number of tokens generated per sample with the RPO method.}
    \label{tab:avg_tokens_RPO_1}
    \renewcommand{\arraystretch}{0.8}
    \setlength{\tabcolsep}{12pt}
    \begin{tabular}{c|cc}
    \toprule
    \textbf{$L$} & \textbf{1.5B Model} & \textbf{7B Model} \\
    \midrule
    300  & 145.88 & 147.06 \\
    500  & 158.41 & 168.17 \\
    800  & 382.20 & 397.89 \\
    \midrule
    GRPO & 2689.51 & 2457.91 \\
    \bottomrule
    \end{tabular}
\end{table}

\begin{table}[t!]
    \centering
    \caption{Training time of RPO and GRPO under 4 epochs with $L = 800$, h represents hours.}
    \label{tab:avg_tokens_RPO_2}
    \renewcommand{\arraystretch}{0.85}
    \setlength{\tabcolsep}{12pt}
    \rowcolors{2}{gray!10}{white}
    \begin{tabular}{c|cc}
    \toprule
    \textbf{Method} & \textbf{1.5B Model} & \textbf{7B Model} \\
    \midrule
    GRPO & 77.28 h  & 84.53 h  \\
    \rowcolor{rpolight}
    +RPO & 8.37 h  & 23.50 h  \\
    \bottomrule
    \end{tabular}
    \vspace{-15pt}
\end{table}

\begin{figure}[t!]
    \centering
    \includegraphics[width=1\linewidth]{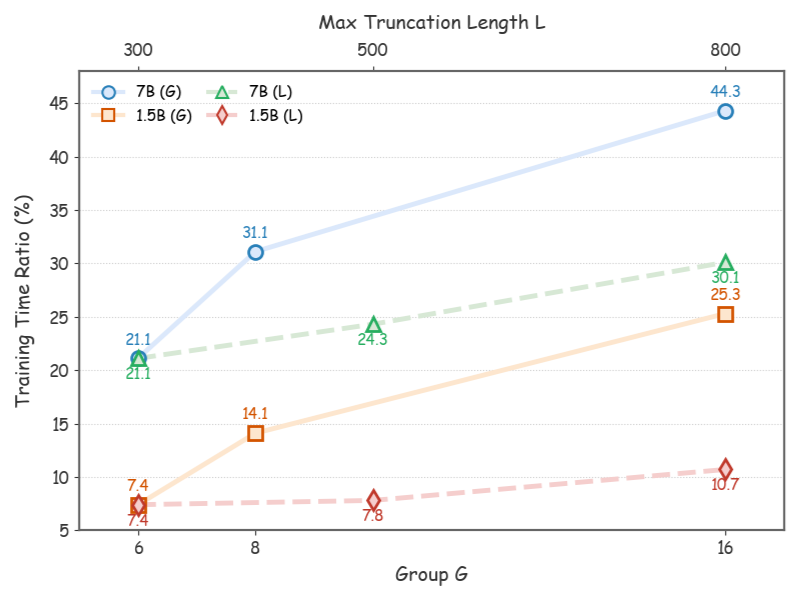}
    \caption{The impact of maximum truncation length $L$ and group size $G$ on the acceleration ratio of POER under the 1.5B and 7B model settings.}
    \label{fig:time_ratio}
    \vspace{-15pt}
\end{figure}

We study training speed for both 1.5B and 7B models. With maximum truncation length fixed at $L=300$, we set per-GPU batch sizes 2 (1.5B) and 1 (7B), and evaluate group sizes $G = 6, 8, 16$. As shown in Figure~\ref{fig:time_ratio}, smaller $G$ yields greater acceleration for RPO, reducing training time to 7.4\% of GRPO for 1.5B and 21.1\% for 7B.  
We also study the effect of $L$ with $G=6$.

The actual training speed is affected by many factors, so we propose a fairer comparison: using the average tokens generated per sample. Since prefill is much faster than decoding, more tokens in prefill lead to shorter decoding time. As shown in Table ~\ref{tab:avg_tokens_RPO_1}, under the original GRPO algorithm, each sample requires an average of 2689.51 tokens and 2457.91 tokens for the 1.5B and 7B models, respectively.

\begin{table*}[ht]
\centering
\setlength{\tabcolsep}{3.8pt}  
\renewcommand{\arraystretch}{0.85}
\caption{Performance of GRPO and RPO on Evaluation Datasets in Multi-Step Iteration Scenarios.}

\begin{tabular}{lccccccr}
\toprule
\textbf{Model} & \textbf{AIME25} & \textbf{AIME24} & \textbf{MATH500} & \textbf{AMC23} & \textbf{Minerva} & \textbf{OlyB}  & \textbf{Avg} \\
\midrule

DeepSeek-R1-Qwen-7B
& 43.3 & 55.5 & 92.8 & 90.0 & 44.5 & 67.4 & 65.6 \\

\quad + GRPO(w/o R) 
& 40.0 & 50.0 & 94.2 & 90.0 & 41.2 & 66.7 & 63.7 \\

\rowcolor{rpolight}
\quad \quad + RPO
& \valdiff{46.6}{40.0} 
& \valdiff{56.7}{50.0} 
& \valdiff{92.8}{94.2} 
& \valdiff{90.0}{90.0} 
& \valdiff{41.4}{41.2} 
& \valdiff{66.1}{66.7} 
& \valdiff{65.6}{63.7} \\

\midrule
DeepSeek-R1-Qwen-1.5B
& 16.7 & 28.8 & 82.2 & 62.9 & 26.5 & 43.3 & 43.4 \\

\quad + GRPO(w/o R) 
& 10.0 & 10.0 & 67.0 & 45.0 & 20.6 & 31.4 & 34.8 \\

\rowcolor{rpolight}
\quad \quad + RPO
& \valdiff{20.0}{10.0} 
& \valdiff{36.7}{10.0} 
& \valdiff{82.8}{67.0} 
& \valdiff{72.5}{45.0} 
& \valdiff{29.4}{20.6} 
& \valdiff{51.5}{31.4} 
& \valdiff{48.8}{34.8} \\

\bottomrule
\end{tabular}
\label{tab:acc stability}
\end{table*}

\begin{figure*}[htbp]
  \centering
  \begin{minipage}{0.46\textwidth}
    \centering
    \includegraphics[width=\linewidth]{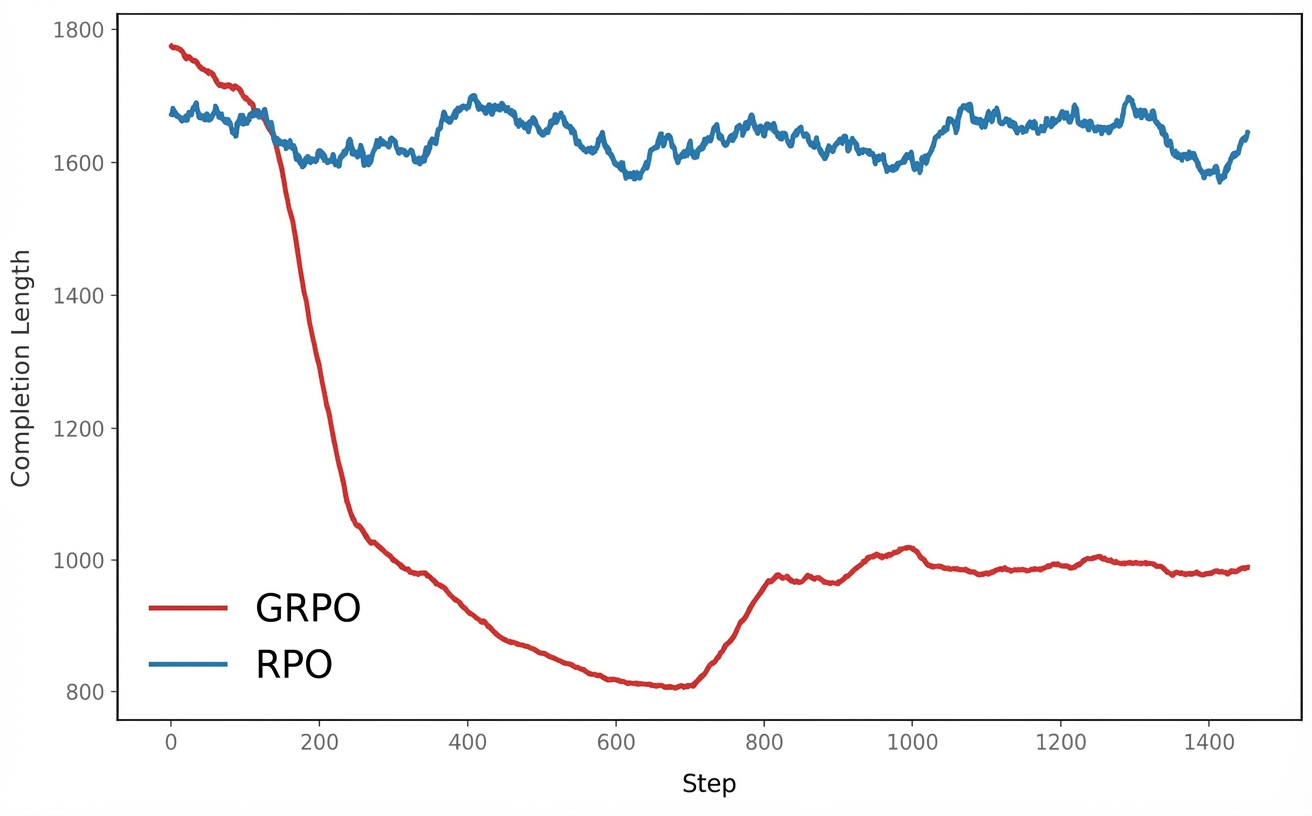} 
    \caption{Response length of RPO and GRPO with full-trajectory optimization on 1.5B model across training steps.}
    \label{more_step_smooth_1.5b}
  \end{minipage}
  \hfill 
  \begin{minipage}{0.48\textwidth}
    \centering
    \includegraphics[width=\linewidth]{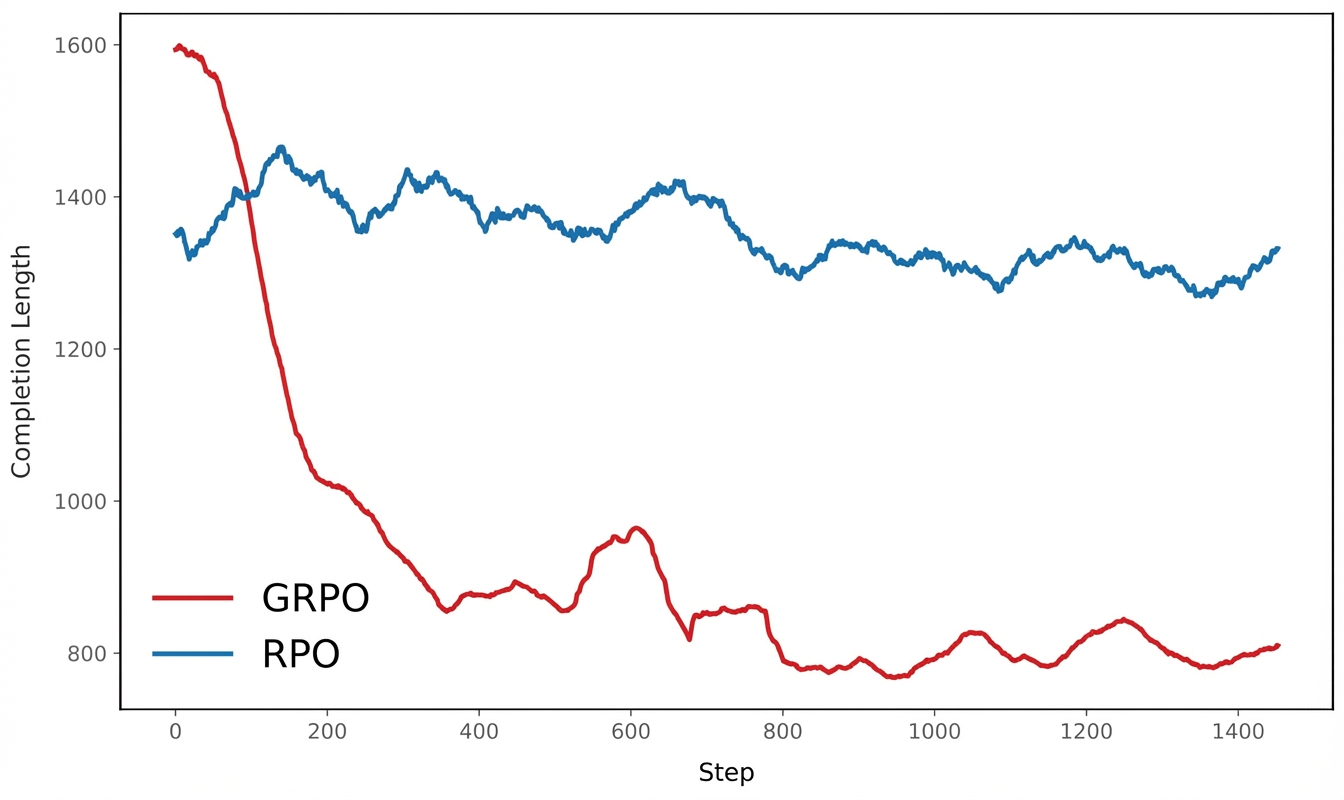} 
   \caption{Response length of RPO and GRPO with full-trajectory optimization on 7B model across training steps.}
    \label{more_step_smooth_single_y-7b}
  \end{minipage}
\vspace{-15pt}
\end{figure*}

In contrast, with the RPO algorithm, the number can be reduced to as low as 145.88 tokens and 147.06 tokens. From the perspective of the decode stage, the time overhead of RPO is only about 5\% of that of GRPO. Table~\ref{tab:avg_tokens_RPO_2} presents the detailed training time overhead of the original GRPO and RPO algorithms over 4 epochs.

It is worth noting that the average number of tokens generated by the GRPO algorithm for the 1.5B model is higher than that for the 7B model. However, when constrained by RPO, the number of tokens generated is lower. This is because the RPO algorithm preserves shorter correct answers, and the exploration capability of the 1.5B model, once guided, is weaker compared to that of the 7B model, leading to this phenomenon.

\subsection{Stability Analysis}
Conventional reinforcement learning methods, such as GRPO and PPO, are prone to instability in multi-step training, with performance and response length often deteriorating as iterations increase. Accordingly, GRPO fine-tuning typically limits iteration numbers, using accuracy and response length to measure model degradation\citep{deepseekai2025deepseekr1incentivizingreasoningcapability}. RPO addresses this by using a cache pool mechanism, and we conduct comparative experiments to quantify its improved training stability over GRPO.

During the training process, we use a batch size of 18 to train the 7B and 1.5B models for four epochs, and monitor changes in response length and model performance, as shown in Figure~\ref{more_step_smooth_1.5b} and \ref{more_step_smooth_single_y-7b}. In this multi-step iterative training setup, GRPO experiences a collapse in response length around the 200th iteration, while RPO maintains stable response lengths throughout the process. On the other hand, as shown in Table~\ref{tab:acc stability}, the model performance after training with GRPO deteriorated, especially for the 1.5B model, where accuracy dropped by 8.6\%. In contrast, RPO results in a 5.4\% improvement in accuracy.

Beyond the instability from reward sparsity, GRPO suffers from strong locality due to its inter-group comparison strategy, limiting performance improvements. RPO mitigates this by introducing an experience cache, using an external cached policy $\pi_{\mathcal{C}}$ to approximate the main policy $\pi_\theta$ during updates. This provides a global context, enhances training stability, and allows RPO to maintain consistent performance over long iterations.




\section{More Analysis}

\paragraph{Impact of Maximum Truncation Length and $\alpha$.}Intuitively, the maximum truncation length $L$ and $\alpha$ are not independent factors. To study their effect on training, we train the model with combinations of $L = 300, 0.5\ell, \ell$ and $\alpha = 0, 0.01, 0.1, 1$. Notably, when $\alpha = 0$, the intra-group length-aware reward is disabled, so all correct reasoning paths receive the same reward.  

\begin{figure}[htbp]
    \centering
    \includegraphics[width=0.45\textwidth]{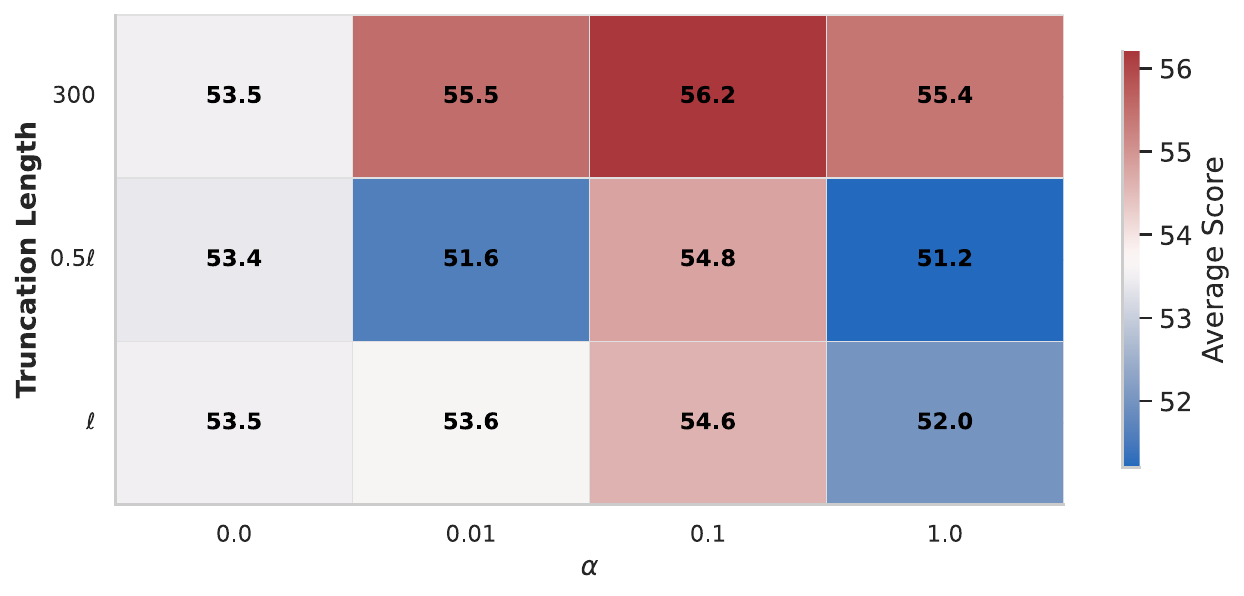}
    \caption{Heatmap of the Impact of $L$ and $\alpha$ on Model Accuracy.}
    \label{fig:heatmap}
    \vspace{-15pt}
\end{figure}

The DeepSeek-R1-Qwen-1.5B model is trained for two epochs with a batch size of 336 to amplify differences in training outcomes for easier observation. The evaluation results are shown in Figure~\ref{fig:heatmap}: under a fixed $L$, performance first improves as $\alpha$ increases and then declines.

\begin{figure}[htbp]
    \centering
    \includegraphics[width=1\linewidth]{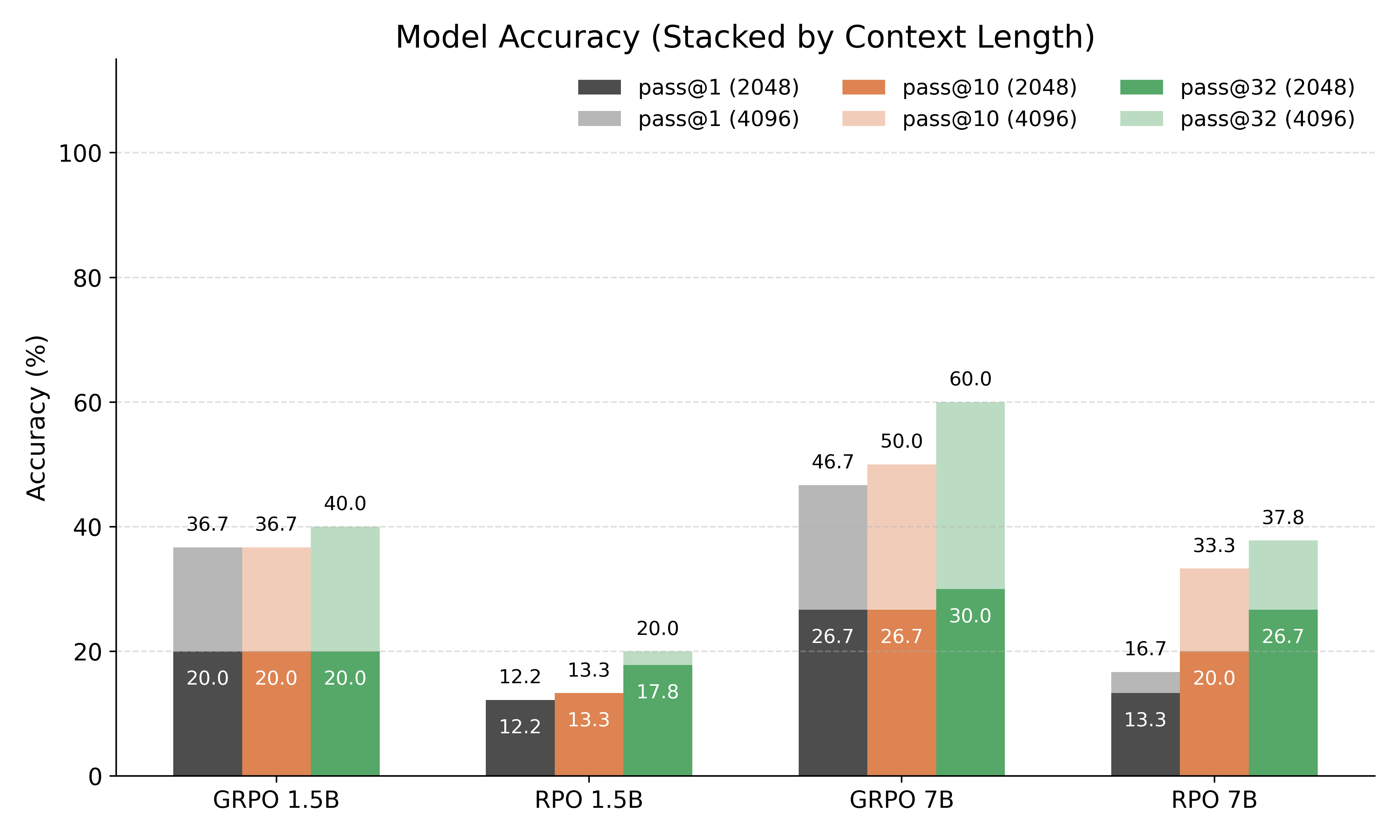}
    \caption{With the maximum truncation length set to 300, the Pass@N performance of the RPO algorithm and the full-path optimization GRPO algorithm on the AIME24 dataset is shown. The opaque bars correspond to the case with a maximum generation length of 2048, while the transparent bars correspond to the case with a maximum generation length of 4096.}
    \label{fig:grpo_poer_2048_4096}
    \vspace{-15pt}
\end{figure}

\paragraph{Effect of max\_length on Exploration Capability.} To investigate the impact of \texttt{max\_length} settings on the model's initial exploration ability, 
we set the maximum truncate length of RPO to 300 and examined the performance of the 1.5B and 7B models 
on the AIME24 dataset under two settings: \texttt{max\_length = 2048} and \texttt{max\_length = 4096}. 
As shown in Figure~\ref{fig:grpo_poer_2048_4096}, RPO demonstrates lower exploration ability compared to GRPO, 
and the gap between the two methods gradually widens as \texttt{max\_length} increases. 
This result also indicates that RPO exhibits a certain disadvantage in exploration ability 
during the early iterations. However, this result compares exploration capabilities without updating the cache pool. In this scenario, RPO’s exploration ability is clearly inferior to GRPO, which adopts a full-path optimization strategy.

\begin{figure}[htbp]
    \centering
    \includegraphics[width=1\linewidth]{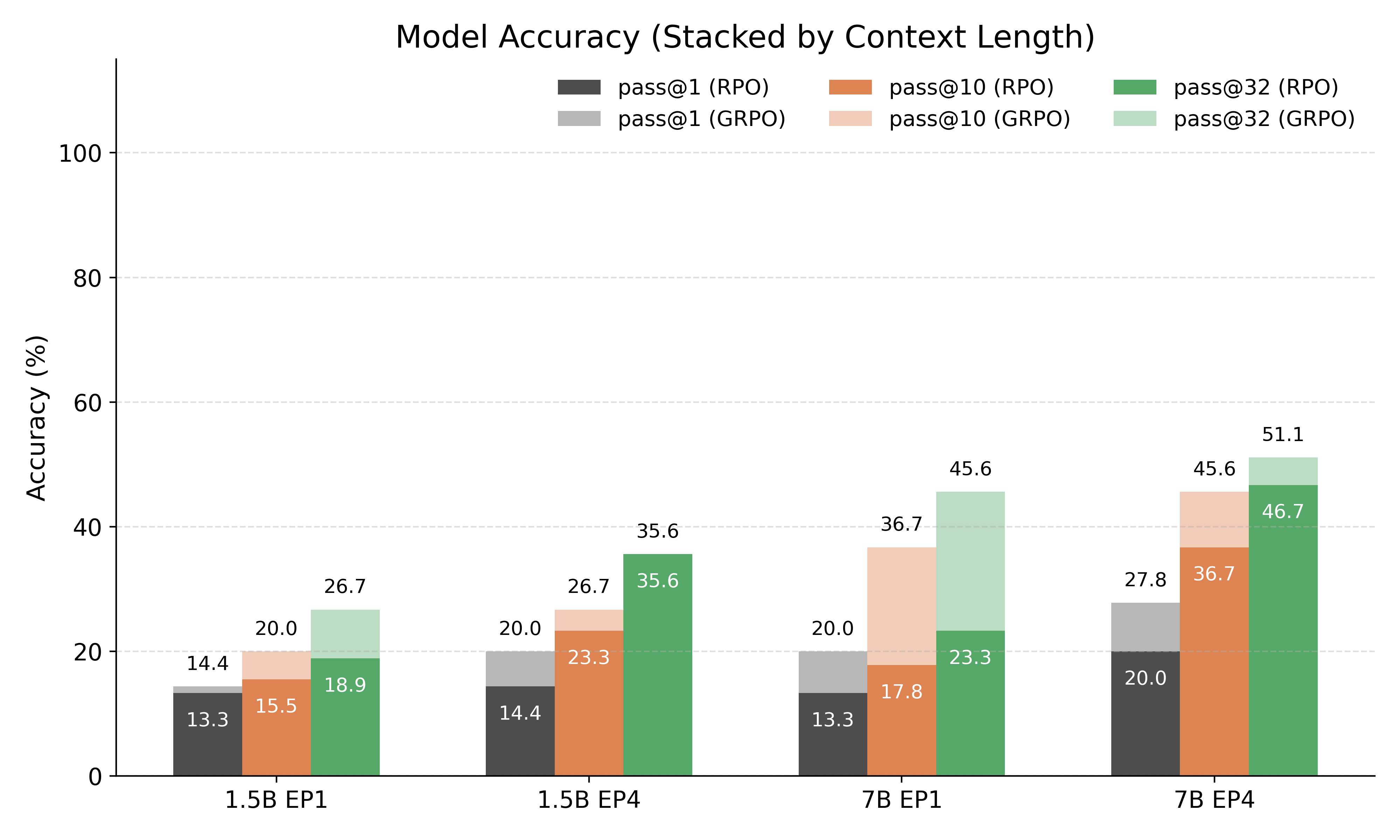}
    \caption{With the maximum truncation length set to 300 and the maximum generation length set to 4096, we investigate the Pass@N performance of the RPO algorithm and the full-path optimization GRPO algorithm on the AIME24 dataset under training epochs set to 1 and 4. The transparent bars represent GRPO, while the opaque bars represent RPO.}
    \label{fig:pass_n_epoch}
    \vspace{-25pt}
\end{figure}

\paragraph{Impact of Cache Pool Update Strategy on Model's Pass@N Performance.} To study the effect of training epochs on exploration capability, we evaluate the 1.5B and 7B models on the AIME24 dataset with \texttt{max\_length} set to 4096 under \texttt{epoch = 1,4}. As shown in Figure~\ref{fig:pass_n_epoch}, as the number of training epochs increases and the cache pool is updated, the model is able to explore more diverse and higher-quality solution paths, with performance steadily improving to eventually match or become comparable to the full-path optimization GRPO.

Overall, although RPO exhibits reduced raw exploration ability in the early stages, the experience cache and the epsilon-greedy strategy guide the model toward higher-quality paths, enabling RPO’s exploration capability in later stages to be essentially on par with that of full-path GRPO. As a trade-off strategy, RPO sacrifices some early exploration capacity but achieves significant training acceleration while ultimately attaining comparable final performance, making this trade-off clearly worthwhile.

\section{Conclusion}
In this paper, we propose RPO, a plug-and-play algorithm for optimizing reinforcement fine-tuning of large language models. RPO introduces an experience replay mechanism that enables the model to leverage previously collected high-quality responses during generation. This approach substantially reduces training time while improving model performance and enhancing stability during the reinforcement fine-tuning process.

\section*{Limitations}
The main limitation of the RPO algorithm is that it sacrifices response diversity for training speed. By sharing historical reasoning prefixes, the generated results within a group tend to converge, which reduces the variance of reward signals and necessitates a reliance on additional "length-aware reward shaping" to maintain performance.



\bibliography{custom}

\onecolumn
\setcounter{equation}{0} 
\renewcommand{\theequation}{A-\arabic{equation}} 
\appendix

\setcounter{equation}{0} 
\renewcommand{\theequation}{B-\arabic{equation}} 
\section{Proof of RPO Gradient Stability}
\label{sec:proof}

 \paragraph{Policy gradient estimation.} The reasoning process of traditional full-path reasoning optimization and \emph{Replay-based Policy Optimization}(RPO) can be expressed as $\pi(\cdot|q_k)$ and $\pi(\cdot|(q_k,a_k))$, where $q_k \in C_{raw}$, is a sample in raw training dataset $C_{raw}$ and corresponding $(q_k, a_k) \in C$, is one example in training replay buffer $C$, which updates during the training process.

For any sample $q_k$, it holds that $(q_k) \subset (q_k,a_k)$.  
Hence, for the response space of an arbitrary policy model, the total variance can give  
\begin{equation}
    \operatorname{Var}(\,\cdot \mid q_k)
      =\mathbb{E}_{(q_k,a_k)\mid q_k}
        \!\Bigl[\operatorname{Var}\bigl(\,\cdot \mid (q_k,a_k)\bigr)\Bigr]
      +\operatorname{Var}_{(q_k,a_k)\mid q_k}
        \!\Bigl(\mathbb{E}\bigl[\,\cdot \mid (q_k,a_k)\bigr]\Bigr).
\end{equation}
Because the second term on the right‐hand side is non‑negative, i.e. $\text{Var}_{(q_k,a_k)|q_k}\left(\mathbb{E}[\cdot |(q_k,a_k)]\right) \geq 0 $, we obtain
\begin{equation}
\operatorname{Var}(\,\cdot \mid q_k)\;
      \ge\;
      \mathbb{E}_{(q_k,a_k)\mid q_k}
        \!\Bigl[\operatorname{Var}\bigl(\,\cdot \mid (q_k,a_k)\bigr)\Bigr].
\end{equation}
Treating $(q_k,a_k)$ as an augmentation of $q_k$ allows this inequality to simplify to  
\begin{equation}
\label{eq:variance_comparison}
\operatorname{Var}(\,\cdot \mid q_k)
      \;\ge\;
      \operatorname{Var}\bigl(\,\cdot \mid (q_k,a_k)\bigr).
\end{equation}
In the policy space, \eqref{eq:variance_comparison} becomes
\begin{equation}
\operatorname{Var}\bigl(\pi_\theta(a_g\mid q_k)\bigr)
      \;\ge\;
\operatorname{Var}\bigl(\pi_\theta(a_g\mid(q_k,a_k))\bigr).
\end{equation}

Assume there exists a parameter vector $\theta_0$ such that the policy can be locally approximated by the first‑order expansion
\begin{equation}
\pi_\theta
      =\pi_{\theta_0}
       +\nabla\!\pi^{\!\top}(\theta_0)\bigl(\theta-\theta_0\bigr).
\end{equation}
The variance of~$\pi_\theta$ in a neighbourhood of~$\theta_0$ can then be estimated as
\begin{equation}
\sigma^2\! \bigl(\pi_\theta\bigr)
      \approx      \nabla\!\pi_{\theta_0}^{\!\top}\Sigma_\theta\,\nabla\!\pi_{\theta_0},
\end{equation}
where $\Sigma_\theta$ denotes the covariance matrix of the parameter estimates.

Throughout training, the realizations of $\pi_\theta$ can be treated as i.i.d.\ random variables.
As the sample size $n\to\infty$, the empirical mean and variance converge to
\[
\mu\!\bigl(\pi_\theta\bigr)
      =\mathbb{E}\!\bigl[\pi_\theta\bigr]
      =\frac1n\sum_{i=1}^n(\pi_\theta)_i,
\qquad
\operatorname{Var}\!\bigl(\pi_\theta\bigr)
      =\mathbb{E}\!\bigl[\pi_\theta^2\bigr]
       -\mu\!\bigl(\pi_\theta\bigr)^2
      \approx
      \frac{n}{n-1}\,
      \nabla\!\pi_{\theta_0}^{\!\top}\Sigma_\theta\,\nabla\!\pi_{\theta_0}.
\]

On account of
$\mathbb{E}[\pi_\theta^2]=\mu(\pi_\theta)^2$, while $\Sigma_\theta=I$ also holds, then
\begin{equation}
\sigma^2\!\bigl(\pi_\theta\bigr)
      =\operatorname{Var}\!\bigl(\pi_\theta\bigr)
      =\nabla\!\pi_{\theta_0}^{\!\top}\nabla\!\pi_{\theta_0}
      =\bigl \| \nabla\!\pi_{\theta_0}\bigr\|_2^{\,2}.
\end{equation}
Combining the above with \eqref{eq:variance_comparison} yields the policy space gradient estimation.
\begin{equation}
\label{eq:core_english}
\Bigl\lVert\nabla\!\pi_\theta\bigl(a_g\mid q_k\bigr)\Bigr\rVert_2
      \;\ge\;
      \Bigl\lVert\nabla\!\pi_\theta\bigl(a_g\mid(q_k,a_k)\bigr)\Bigr\rVert_2.
\end{equation}
This establishes that conditioning on the augmented information $(q_k,a_k)$ strictly reduces—or at worst preserves—the magnitude of the policy‐gradient variance.

\medskip
\noindent

Let's review the GRPO update policy. At training step~$t$, the optimisation target of \emph{Generative Reinforcement Policy Optimisation} (GRPO) can be written as
\begin{flalign}
\begin{aligned}
&\mathcal{J}_{\mathrm{GRPO}}(\theta_t) =
\mathbb{E}_{(q,a)\sim \mathcal{C}^{(t-1)}(Q,A),\{o_i\}_{i=1}^G \sim \pi_{\theta_{t-1}}(\cdot | q))}  \\
& \quad \Bigg[\frac{1}{G} \sum_{i=1}^G \frac{1}{|o_i|} \sum^{|o_i|}_{j=1} \min \bigg(
    r_{i,j}(\theta_t) \hat{A}_{i,j},
    \mathrm{clip}\Big( r_{i,j}(\theta_t), 1-\epsilon, 1+\epsilon \Big) \hat{A}_{i,j} - \beta D_{\mathrm{KL}}(\pi_{\theta_t} \| \pi_{\mathrm{ref}})
\bigg) \Bigg]
\end{aligned}
\end{flalign}
Here
\[
r_{i,j}(\theta_t)=
      \frac{\pi_{\theta_t}(o_{i,j}\mid q)}%
           {\pi_{\theta_{t-1}}(o_{i,j}\mid q)},
\qquad
\hat A_{i,j}=
      \frac{R_i-\operatorname{mean}\bigl(\{R_i\}_{i=1}^{G}\bigr)}%
           {\operatorname{std}\bigl(\{R_i\}_{i=1}^{G}\bigr)}.
\]

Relative to GRPO, \textsc{RPO} only changes the policy ratio by conditioning on the prefix~$o_{i,<j}$:
\begin{equation}
r^{\mathrm{RPO}}_{i,j}(\theta_t)
      =\frac{\pi_{\theta_t}(o_{i,j}\mid q,\,o_{i,<j})}%
             {\pi_{\theta_{t-1}}(o_{i,j}\mid q,\,o_{i,<j})}.
\end{equation}

Because the \texttt{clip} operation truncates high‑error updates, both algorithms behave identically whenever clipping is activated.
\paragraph{Gradient of the optimization policy.}
For the plain GRPO ratio one obtains
\begin{equation} 
\nabla_\theta\,r_{i,j}(\theta_t)
      =\frac{\nabla_\theta\pi_{\theta_t}(o_{i,j}\mid q)}%
             {\pi_{\theta_{t-1}}(o_{i,j}\mid q)}
      =\frac{\pi_{\theta_t}(o_{i,j}\mid q)}%
             {\pi_{\theta_{t-1}}(o_{i,j}\mid q)}
       \,\nabla_\theta\!\log\pi_{\theta_t}(o_{i,j}\mid q).
\end{equation}
The Kullback–Leibler divergence with respect to a frozen reference policy $\pi_{\mathrm{ref}}$ satisfies
\begin{equation}
\begin{aligned}
\nabla_\theta D_{\mathrm{KL}}\!\bigl(\pi_{\theta_t}\,\Vert\,\pi_{\mathrm{ref}}\bigr)
   &=\nabla_\theta\mathbb{E}_{\pi_{\theta_t}}
        \Bigl[\log \pi_{\theta_t}-\log\pi_{\mathrm{ref}}\Bigr] \\
   &=\mathbb{E}_{\pi_{\theta_t}}
        \Bigl[
             \nabla_\theta\log \pi_{\theta_t}
             +\bigl(\log \pi_{\theta_t}-\log\pi_{\mathrm{ref}}\bigr)
              \nabla_\theta\log \pi_{\theta_t}
        \Bigr] \\
   &=\mathbb{E}_{\pi_{\theta_t}}
        \Bigl[
             \nabla_\theta\log \pi_{\theta_t}
             \,\bigl(\log\tfrac{\pi_{\theta_t}}{\pi_{\mathrm{ref}}}+1\bigr)
        \Bigr]
\end{aligned}
\end{equation}

As $\mathbb{E}_{\pi_{\theta_t}}[\nabla_\theta\log\pi_{\theta_t}]=0$ by normalisation, the expression simplifies to
\begin{equation}
\begin{aligned}
 \nabla_\theta D_{\text{KL}} =& \mathbb{E}_{\pi_\theta} \left[ \nabla_\theta \log \pi_\theta(o|s) \cdot \log {\pi_\theta(o|s)} \right] \\
 =&\sum_{\pi_\theta}\pi_\theta \nabla_\theta \log \pi_\theta(o|s) \cdot \log {\pi_\theta(o|s)} \\
 =&\frac{1}{|o_i|}\sum_{t=1}^{|o_i|}\pi_\theta \nabla_\theta \log \pi_\theta(o_i|s) \cdot \log {\pi_\theta(o_i|s)}
\end{aligned}
\end{equation}

\paragraph{Resulting policy gradient.}
Aggregating the intra‑group updates yields the estimator
\begin{equation}
\begin{aligned}
\nabla_\theta \mathcal{J}_{\mathrm{ALL}}
 &=\mathbb{E}_{q,\{o_i\}}
     \Biggl[
        \frac{1}{G}\sum_{i=1}^{G}\frac{1}{|o_i|}\sum_{j=1}^{|o_i|}
        \Bigl(
           \tfrac{\hat A_{i,j}}{\pi_{\theta_{t-1}}(o\mid q)}
           -\beta\,\log \pi_{\theta_t}(o\mid q)
        \Bigr)
        \,\pi_{\theta_t}(o\mid q)\,
        \nabla_\theta\!\log \pi_{\theta_t}(o\mid q)
     \Biggr]
\\
 &=\mathbb{E}_{q,\{o_i\}}
     \Biggl[
        \frac{1}{G}\sum_{i=1}^{G}\frac{1}{|o_i|}\sum_{j=1}^{|o_i|}
        \Bigl(
           \tfrac{\hat A_{i,j}}{\pi_{\theta_{t-1}}(o\mid q)}
           -\beta\,\log \pi_{\theta_t}(o\mid q)
        \Bigr)
        \,\nabla_\theta \pi_{\theta_t}(o\mid q)
     \Biggr].
\end{aligned}
\end{equation}

The second line follows by noting that
\(
\pi_{\theta_t}\nabla_\theta\!\log\pi_{\theta_t}
      =\nabla_\theta \pi_{\theta_t}.
\) Equation above provides the final form of the GRPO gradient used for parameter updates at step~$t$.

\medskip
\noindent

\paragraph{Without consideration of KL divergence.}
If the KL--divergence term is temporarily ignored, the GRPO gradient estimator
reduces to
\begin{equation}
\nabla_\theta \mathcal{J}_{\mathrm{ALL}}
   =\mathbb{E}_{q,\{o_i\}}
      \Biggl[
         \frac1G\sum_{i=1}^{G}\frac1{|o_i|}
         \sum_{t=1}^{|o_i|}
         \bigl(\tfrac{\hat A_{i,t}}{\pi_{\theta_{t-1}}}\bigr)\,
         \nabla\pi_\theta(a_g\mid q_k)
      \Biggr].
\end{equation}

Because of equation~\eqref{eq:core_english},
\begin{equation}
\lVert \nabla_\theta \mathcal{J}_{\mathrm{ALL}}\rVert_2
      \;\ge\;
      \lVert\nabla_\theta \mathcal{J}_{\mathrm{RPO}}\rVert_2.
\end{equation}

\paragraph{Including the KL divergence.}
At the initial step ($t=0$) both algorithms share the same reference policy,
hence
\begin{equation}
\nabla_\theta \mathcal{J}_{\mathrm{ALL}}
   =\nabla_\theta \mathcal{J}_{\mathrm{RPO}}.
\end{equation}

For the first update ($t=1$) \eqref{eq:core_english} implies
\begin{equation}
\lVert\nabla\pi_{\theta_1}(a_g\mid q_k)\rVert_2
      \;\ge\;
      \lVert\nabla\pi_{\theta_1}(a_g\mid q_k,a_k)\rVert_2.
\end{equation}
Here we record $\nabla\pi_{\theta_1}(a_g\mid q_k,a_k)$ as $\nabla \pi_{\theta_1}^\prime$.
Consequently, the difference of the two policy gradients becomes
\begin{equation}
\begin{aligned}
\Delta_1
 &\;=\;\|
   \nabla_\theta \mathcal{J}_{\mathrm{ALL}}\|_2
   \;-\;
   \|\nabla_\theta \mathcal{J}_{\mathrm{RPO}}\|_2\\
 &\;=\;
   \mathbb{E}_{q,\{o_i\}}
   \Biggl[
     \frac1G\sum_{i=1}^{G}\frac1{|o_i|}\sum_{t=1}^{|o_i|}
     \Bigl(\|
       (\tfrac{\hat A_{i,1}}{\pi_{\theta_0}}-\beta\log\pi_{\theta_1})
       \nabla\pi_{\theta_1}\|_2
       -
       \|(\tfrac{\hat A_{i,1}^\prime}{\pi_{\theta_0}}-\beta\log\pi_{\theta_1}^\prime)
       \nabla\pi_{\theta_1}^\prime
     \|_2\Bigr)
   \Biggr].
\end{aligned}
\end{equation}

Let 
\(
\delta
   :=\nabla_\theta\pi_{\theta_1}(i)\!-\!\nabla_\theta\pi_{\theta_1}^\prime(i)\ge0.
\)
For one random dimension the mean‑value theorem yields
\begin{equation}
\pi_{\theta_1}(i)\log\pi_{\theta_1}(i)
      -\pi_{\theta_1}^\prime(i)\log\pi_{\theta_1}^\prime(i)
   =\Bigl.\frac{\partial (\pi\log\pi)}{\partial\pi}\Bigr|_{\pi=\zeta}
     (\pi_{\theta_1}(i)-\pi_{\theta_1}^\prime(i)),
\quad
\zeta\in[\pi_{\theta_1}^\prime,\pi_{\theta_1}]\subset[0,1).
\end{equation}
Taking the directional derivative with respect to~$\theta$ gives
\begin{equation}
    \begin{aligned}
        (1+\log \pi_{\theta_1}(i))\nabla \pi_{\theta_1}(i) - (1+\log \pi_{\theta_1}^\prime(i))\nabla \pi_{\theta_1}^\prime(i) =& \frac{\partial \pi_{\theta}\log \pi_{\theta}}{\partial \ \pi_{\theta}}|_{\pi_\theta = \zeta}  (\nabla \pi_{\theta_1}(i) - \nabla \pi_{\theta_1}^\prime(i)) \\
        =&\frac{\partial \pi_{\theta}\log \pi_{\theta}}{\partial \ \pi_{\theta}}|_{\pi_\theta = \zeta} \delta
    \end{aligned}
\end{equation}
Hence
\begin{equation}
\log\pi_{\theta_1}(i)\,\nabla\pi_{\theta_1}(i)
   -\log\pi_{\theta_1}^\prime(i)\,\nabla\pi_{\theta_1}^\prime(i)
   =\bigl(\tfrac{\partial(\pi\log\pi)}{\partial\pi}\bigr|_{\pi=\zeta}-1\bigr)\delta
   =(\log\zeta)\,\delta
   \;\le\;0,
\end{equation}
because $\log\zeta<0$.  

Extending this argument component‑wise to the full
parameter vector shows
\begin{equation}
\log\pi_{\theta_1}\,\nabla\pi_{\theta_1}
    \preceq \log\pi_{\theta_1}^\prime\,\nabla\pi_{\theta_1}^\prime,
\end{equation}
and therefore
\begin{equation}
\|-\beta\log\pi_{\theta_1}\,\nabla\pi_{\theta_1}\|_2
    \geq \| -\beta \log\pi_{\theta_1}^\prime\,\nabla\pi_{\theta_1}^\prime\|_2
   ,
\end{equation}
We have thus established
\begin{equation}\label{eq:J_first}
\Vert\nabla_{\theta_1}\mathcal{J}_{\mathrm{ALL}}\rVert_2
      \;\ge\;
      \lVert\nabla_{\theta_1}\mathcal{J}_{\mathrm{RPO}}\rVert_2.
\end{equation}

Let the generic update rule be
\begin{equation}    
\theta_i=\theta_{i-1}+\eta\,\nabla_\theta\mathcal{J}.
\end{equation}

Then
\begin{equation}
\frac{\nabla\pi_{\theta_i}}{\pi_{\theta_{i-1}}}
   =\frac{\nabla\bigl(\pi_{\theta_{i-1}}
                      +\eta\,\nabla\pi_{\theta_{i-1}}
                       \nabla_{\theta_{i-1}}\mathcal{J}\bigr)}
          {\pi_{\theta_{i-1}}}
   \;\ge\;
   \frac{\nabla\pi_{\theta_{i-1}}}{\pi_{\theta_{i-1}}}
   =\nabla\log\pi_{\theta_{i-1}},
\end{equation}

i.e.\ each step re‑enters the original policy‑gradient (PG) regime.  
Using \eqref{eq:J_first} one obtains for every $i\ge1$
\begin{equation}
\frac{\nabla\pi_{\theta_i}}{\pi_{\theta_{i-1}}}
   -\frac{\nabla\pi_{\theta_i}^\prime}{\pi_{\theta_{i-1}}^\prime}
   \;\ge\;
   \frac{\nabla\pi_{\theta_{i-1}}}{\pi_{\theta_{i-1}}}
   -\frac{\nabla\pi_{\theta_{i-1}}^\prime}{\pi_{\theta_{i-1}}^\prime}
   \;\ge\;
   \nabla\log\frac{\pi_{\theta_{i-1}}}{\pi_{\theta_{i-1}}^\prime}.
\end{equation}
By induction this yields the general relation
\begin{equation}
\label{eq:J_general}
\|\nabla_\theta\mathcal{J}_{\mathrm{ALL}}\|_2
      \;\ge\;
      \|\nabla_\theta\mathcal{J}_{\mathrm{RPO}}\|_2
\qquad
\text{for all optimisation steps.}
\end{equation}
Equation~\eqref{eq:J_general} completes the proof that, under identical
hyper‑parameters, GRPO provides gradient updates at least as large as those
of RPO, both without and with the KL divergence.
Meanwhile, since both of them follow a normal distribution with zero mean, it follows that:
\begin{equation}
Var(\|\nabla_{\theta}\mathcal{J}_{\mathrm{RPO}}\|_2)\leq Var(\|\nabla_{\theta} \mathcal{J}_{\mathrm{ALL}}\|_2)
\end{equation}

\setcounter{equation}{0} 
\renewcommand{\theequation}{C-\arabic{equation}} 
\section{theorem}
\label{proof:2}

\paragraph{Preliminary.}
Let $q$ denote the prompt, $o$ a sampled response with length $\ell=\mathrm{len}(o)$, and let the group-wise reference length be $\ell_{\mathrm{ref}}=\tfrac{1}{G}\sum_{i=1}^G \mathrm{len}(o_i)$. Write $\Delta\ell=\ell-\ell_{\mathrm{ref}}$ and fix a window $|\Delta\ell|\le \tau$. For a sensitivity parameter $\alpha>0$ define the length weight $s_\alpha(\ell)=\sigma\!\big(\alpha(\ell_{\mathrm{ref}}-\ell)\big)=(1+e^{-\alpha(\ell_{\mathrm{ref}}-\ell)})^{-1}\in(0,1)$ and the shaped reward $R_\alpha=\mathrm{clip}\!\big(s_\alpha(\ell)\,r,\;m,\;\mathcal M\big)$ with clipping bounds $m<\mathcal M$, where $r$ is the original per-sample reward. 

Consider RPO with replay distribution $\mu$ and current policy $\pi_\theta$, truncated importance ratio $\rho=\min\!\big(c,\tfrac{\pi_\theta(o\mid q)}{\mu(o\mid q)}\big)$ for a constant $c\ge 1$, token-averaged score function $\nabla_\theta \log \pi_\theta(o\mid q)=\tfrac{1}{|o|}\sum_{j=1}^{|o|}\nabla_\theta \log \pi_\theta(o_j\mid o_{<j},q)$, and a centered advantage $A'_\alpha=R_\alpha-b_\alpha$ with group baseline $b_\alpha=\mathbb{E}[R_\alpha\mid q,\mathrm{group}]$. The single-sample gradient contribution is
\begin{equation}
g_\alpha \;=\; \rho \,\Big(A'_\alpha - \beta \log \pi_\theta(o\mid q)\Big)\,\nabla_\theta \log \pi_\theta(o\mid q).
\label{eq:g-alpha}
\end{equation}
Assume $\|\nabla_\theta \log \pi_\theta(o\mid q)\|\le L$, $\mathbb{E}[r^2]<\infty$, and that 
\begin{enumerate}

    \item the conditional variance $\sigma_A^2(\ell):=\mathrm{Var}(A\mid \ell)$ of the unshaped advantage $A$ is nondecreasing in $\ell$,
    \item the tail probability $\mathbb{P}\!\big(\tfrac{\pi_\theta}{\mu}>c\mid \ell\big)$ is nondecreasing in $\ell$.
\end{enumerate}
If $\alpha\tau\le 1$, then there exists $\alpha^\star>0$ such that for all $0<\alpha\le \alpha^\star$ the mean-squared error $\mathrm{MSE}_\alpha:=\mathrm{Var}(g_\alpha)+\|\mathbb{E}[g_\alpha]-\nabla_\theta J\|^2$ of the RPO gradient estimator with length-aware shaping satisfies
\begin{equation}
\mathrm{MSE}_\alpha
\;<\;
\min\!\Big\{\mathrm{MSE}^{\mathrm{RPO}}_{0},\;\mathrm{MSE}^{\mathrm{ALL}}_{0},\;\inf_{\tilde\alpha>0}\mathrm{MSE}^{\mathrm{ALL}}_{\tilde\alpha}\Big\},
\end{equation}
that is, it strictly improves upon both the unshaped RPO baseline and the ALL baselines in a nontrivial neighborhood of $\alpha=0$.

\paragraph{Proof.}
The proof makes explicit the first-order behavior in $\alpha$ of both the variance and the bias terms. Throughout the window $|\Delta\ell|\le \tau$ the sigmoid admits the uniform Taylor expansion
\begin{equation}
s_\alpha(\ell)
\;=\;
\frac{1}{2}\;-\;\frac{\alpha}{4}\,\Delta\ell\;+\;R_2(\alpha,\ell),
\qquad
|R_2(\alpha,\ell)|\;\le\;C_2\,\alpha^2\tau^2,
\label{eq:s-expansion}
\end{equation}
for some constant $C_2$ independent of $\alpha$ and $\ell$. Writing $R_\alpha=s_\alpha(\ell)\,r$ on the non-clipping region and absorbing the clipping into the moment bounds later, the centered advantage becomes
\begin{equation}
A'_\alpha
\;=\;
\Big(\tfrac{1}{2}\,r - \mathbb{E}[\tfrac{1}{2}\,r]\Big)
\;-\;\frac{\alpha}{4}\Big(\Delta\ell\,r - \mathbb{E}[\Delta\ell\,r]\Big)
\;+\;\underbrace{R_2(\alpha,\ell)\,r - \mathbb{E}[R_2(\alpha,\ell)\,r]}_{=:E_2(\alpha)}.
\label{eq:Aprime-expansion}
\end{equation}
Substituting \eqref{eq:Aprime-expansion} into \eqref{eq:g-alpha} and taking expectations yields
\begin{equation}
\mathbb{E}[g_\alpha] - \mathbb{E}[g_0]
\;=\;
-\frac{\alpha}{4}\;\mathbb{E}\!\Big[\rho\;\big(\Delta\ell\,r - \mathbb{E}[\Delta\ell\,r]\big)\;\nabla_\theta \log \pi_\theta(o\mid q)\Big]
\;+\;\mathbb{E}\!\Big[\rho\;E_2(\alpha)\;\nabla_\theta \log \pi_\theta(o\mid q)\Big].
\label{eq:mean-diff}
\end{equation}
By Cauchy--Schwarz and the bounds on $\rho$ and the score function, the norm of the first term on the right-hand side satisfies
\begin{equation}
\Big\|
\mathbb{E}\!\Big[\rho\;\big(\Delta\ell\,r - \mathbb{E}[\Delta\ell\,r]\big)\;\nabla_\theta \log \pi_\theta\Big]
\Big\|
\;\le\;
c\,L\;\Big(\mathbb{E}\big[(\Delta\ell\,r - \mathbb{E}[\Delta\ell\,r])^2\big]\Big)^{1/2}
\;\le\;
c\,L\,\tau\,\big(\mathbb{E}[r^2]\big)^{1/2},
\end{equation}
hence $\|\mathbb{E}[g_\alpha]-\mathbb{E}[g_0]\|\le \tfrac{\alpha}{4}\,c\,L\,\tau\,(\mathbb{E}[r^2])^{1/2} + c\,L\,\mathbb{E}[|E_2(\alpha)|]$. Using $|E_2(\alpha)|\le 2 C_2 \alpha^2\tau^2 |r|$ and $\mathbb{E}[r^2]<\infty$ gives the bias bound
\begin{equation}
\|\mathbb{E}[g_\alpha]-\mathbb{E}[g_0]\|
\;\le\;
C_b\,\alpha\,\tau \;+\; C_b'\,\alpha^2\tau^2,
\label{eq:bias-bound}
\end{equation}
for constants $C_b,C_b'$ depending only on $(c,L,\mathbb{E}[r^2],C_2)$. Consequently the squared-bias contribution to $\mathrm{MSE}_\alpha$ is $O(\alpha^2\tau^2)$.

For the variance term, expand the second moment as
\begin{equation}
\mathbb{E}\big[\|g_\alpha\|^2\big]
\;\le\;
c^2\,\mathbb{E}\!\Big[\big(A'_\alpha - \beta \log \pi_\theta\big)^2\;\big\|\nabla_\theta \log \pi_\theta\big\|^2\Big]
\;\le\;
c^2 L^2\,\mathbb{E}\!\Big[\big(A'_\alpha - \beta \log \pi_\theta\big)^2\Big].
\end{equation}
The cross terms between $A'_\alpha$ and $\beta \log \pi_\theta$ are uniformly bounded in $\alpha$ by Jensen and the finite second moments of $r$ and $\log\pi_\theta$. The $\alpha$-dependent leading component arises from $\mathbb{E}[{A'_\alpha}^2]$. Within the non-clipping region and after centering, the contribution that depends on length is proportional to
\begin{equation}
\mathbb{E}\!\big[s_\alpha(\ell)^2\,\sigma_A^2(\ell)\big]
\;=\;
\mathbb{E}[s_\alpha(\ell)^2]\;\mathbb{E}[\sigma_A^2(\ell)]
\;-\;
\mathrm{Cov}\!\big(s_\alpha(\ell)^2,\;\sigma_A^2(\ell)\big).
\label{eq:reverse-cheb}
\end{equation}
Since $s_\alpha(\ell)$ is nonincreasing in $\ell$ while $\sigma_A^2(\ell)$ is nondecreasing in $\ell$ by assumption, the reverse Chebyshev inequality ensures that the covariance in \eqref{eq:reverse-cheb} is nonpositive and is strictly negative unless $s_\alpha(\ell)^2$ and $\sigma_A^2(\ell)$ are almost surely constant. Differentiating $\mathbb{E}[s_\alpha(\ell)^2]$ at $\alpha=0$ and using \eqref{eq:s-expansion} yields $\mathbb{E}[s_\alpha(\ell)^2]=\tfrac{1}{4}+O(\alpha^2\tau^2)$, while differentiating the covariance at $\alpha=0$ gives a strictly negative slope whenever the variance $\sigma_A^2(\ell)$ is not degenerate. Therefore there exists $\eta>0$ such that
\begin{equation}
\mathrm{Var}(g_\alpha)
\;\le\;
c^2 L^2\Big(\tfrac{1}{4}\,\overline{\sigma_A^2}\;-\;\eta\,\alpha\;+\;O(\alpha^2\tau^2)\Big)\;+\;C_{\beta},
\label{eq:var-bound}
\end{equation}
where $\overline{\sigma_A^2}=\mathbb{E}[\sigma_A^2(\ell)]$ and $C_\beta$ collects the $\beta$-dependent but $\alpha$-independent finite terms.

The RPO-specific truncation bias can be written as the deviation between the untruncated importance-weight estimator and the truncated one. Let $w=\tfrac{\pi_\theta}{\mu}$ and $X_\alpha=(A'_\alpha-\beta \log \pi_\theta)\,\nabla_\theta \log \pi_\theta$. The bias vector equals
\begin{equation}
b_{\mathrm{clip}}(\alpha)
\;=\;
\mathbb{E}\big[(w-\rho)\,X_\alpha\big]
\;=\;
\mathbb{E}\big[(w-c)^+\,X_\alpha\big],
\label{eq:clip-bias}
\end{equation}
so that $\|b_{\mathrm{clip}}(\alpha)\|\le \mathbb{E}\big[(w-c)^+\,\|X_\alpha\|\big] \le \mathbb{E}\big[(w-c)^+\,(|A'_\alpha|+|\beta||\log\pi_\theta|)\,L\big]$. 

Assumption 2 implies that the event $\{w>c\}$ is more likely at larger $\ell$, whereas $|A'_\alpha|$ is reduced at larger $\ell$ because $s_\alpha(\ell)$ decreases with $\ell$ and the clipping of $R_\alpha$ further upper-bounds its magnitude. 

Consequently the mapping $\alpha\mapsto \|b_{\mathrm{clip}}(\alpha)\|$ is nonincreasing for small $\alpha$, and in particular $\|b_{\mathrm{clip}}(\alpha)\|\le \|b_{\mathrm{clip}}(0)\|$. Since $\mathrm{MSE}_\alpha$ contains $\|b_{\mathrm{clip}}(\alpha)\|^2$, this term does not increase with $\alpha$ near zero.

Combining \eqref{eq:bias-bound} and \eqref{eq:var-bound} and adding the nonincreasing truncation-bias square gives
\begin{equation}
\mathrm{MSE}_\alpha
\;=\;
\mathrm{Var}(g_\alpha)\;+\;\big\|\mathbb{E}[g_\alpha]-\nabla_\theta J\big\|^2
\;\le\;
c^2 L^2\Big(\tfrac{1}{4}\,\overline{\sigma_A^2}\;-\;\eta\,\alpha\;+\;O(\alpha^2\tau^2)\Big)
\;+\;\|b_{\mathrm{clip}}(\alpha)\|^2
\;+\;O(\alpha^2\tau^2).
\label{eq:mse-master}
\end{equation}
Choosing $\alpha^\star>0$ sufficiently small so that the linear decrease $-\eta\,\alpha$ dominates the aggregated $O(\alpha^2\tau^2)$ remainders ensures that $\mathrm{MSE}_\alpha<\mathrm{MSE}_0^{\mathrm{RPO}}$ for all $0<\alpha\le \alpha^\star$ with $\alpha\tau\le 1$. 

Since GRPO coincides with the on-policy case without any truncation channel, its $\alpha$-dependence shares the same variance reduction mechanism but lacks the nonincreasing truncation-bias term $\|b_{\mathrm{clip}}(\alpha)\|^2$; therefore the same choice of $\alpha$ also yields $\mathrm{MSE}_\alpha^{\mathrm{RPO}}<\min\{\mathrm{MSE}_0^{\mathrm{ALL}},\inf_{\tilde\alpha>0}\mathrm{MSE}_{\tilde\alpha}^{\mathrm{ALL}}\}$ whenever $\|b_{\mathrm{clip}}(0)\|>0$, which holds generically under assumption (ii). This proves the stated improvement.

\paragraph{Remark.}
The token-wise averaging in GRPO, $\tfrac{1}{|o|}\sum_{j=1}^{|o|}$, multiplies the effective per-sample weight by $|o|^{-1}$ and thus accentuates the negative covariance in \eqref{eq:reverse-cheb}, because $|o|^{-1}$ is also nonincreasing in $\ell$. The group baseline $b_\alpha$ used to define $A'_\alpha$ guarantees that the constant component of $s_\alpha(\ell)$ is removed, while the window condition $\alpha\tau\le 1$ keeps $s_\alpha(\ell)$ within the near-linear regime where \eqref{eq:s-expansion} is valid and the remainder terms are uniformly controlled.

\twocolumn
\section{Using a large model's cache pool to guide small model training}
We design the following experiment to explore whether introducing a more powerful model for question sampling during the cache pool initialization phase can influence the resulting cache policy, thereby allowing the original model to indirectly benefit from the distillation of the stronger model's reasoning capabilities.

Specifically, we use the cache pool initialized by deepseek-r1-qwen-7b as the initial cache pool for deepseek-r1-qwen-1.5b. Then, following the original experimental setup, we train for two epochs and evaluate the final performance.
As shown in Table~\ref{fig:7b on 1b}, when trained using the cache pool generated by the 7B model, the 1.5B model did not significantly improve performance.

\begin{table}[H]
\centering
\small  
\setlength{\tabcolsep}{6pt}  
\renewcommand{\arraystretch}{1.1} 
\caption{The Performance of a 7B Model's Cache Pool on a 1.5B Model.}
\resizebox{\linewidth}{!}{
\begin{tabular}{l@{\hskip 3pt}c@{\hskip 4pt}c@{\hskip 4pt}c@{\hskip 4pt}c@{\hskip 4pt}c@{\hskip 4pt}c@{\hskip 10pt}c}
\toprule
\textbf{Model} & \textbf{AIME24} & \textbf{MATH500} & \textbf{AMC23} & \textbf{Minerva} & \textbf{OlyB}  & \textbf{Avg} \\
\midrule
\multirow{1}{*}{DS-R1-Qwen-1.5B}
& 23.3 & 84.8 & 75.0 & 28.7 & 53.5 & 53.1 \\
\bottomrule 
\label{fig:7b on 1b}
\end{tabular}
}
\end{table}

\section{Cost Overhead}
\label{sec:cost overhead}
In this section, we present the cost overhead of several additional open-source models with the same parameters, as well as that of the series of models based on our RPO algorithm.

\begin{table}[H]
\centering
\caption{Comparison of data usage and computational costs with 1.5B models.}
\resizebox{\linewidth}{!}{
    \begin{tabular}{l|c|c|c}
    \toprule
     & \textbf{DeepScaleR-1.5B} & \textbf{Still-3-1.5B} & \textbf{RPO} \\
    \midrule
    \textbf{Base Model} & \multicolumn{3}{c}{DeepSeek-R1-Distill-Qwen-1.5B} \\
    \midrule
    Hardware & 8$\times$ A100 80GB & 1$\times$8 A100 80GB & 1$\times$8 A100 80GB \\
    Time & 240h & 150h & 3h \\
    \midrule
    \textbf{Cost Est.} & \$3629 & \$2268 & \$24 \\
    \bottomrule
    \end{tabular}
}
\end{table}

\begin{table}[H]
\centering
\caption{Comparison of data usage and computational costs with 7B models.}
\resizebox{\linewidth}{!}{
    \begin{tabular}{l|c|c}
    \toprule
     & \textbf{rStar-Math-7B\citep{guan2025rstarmathsmallllmsmaster}} & \textbf{Eurus-2-7B-PRIME}  \\
    \midrule
    \textbf{Base Model}  & \multicolumn{2}{c}{Qwen2.5-Math-7B} \\
    \midrule
    Hardware & \begin{tabular}[c]{@{}c@{}}10$\times$8 H100 80GB,\\ 15$\times$4 A100 40GB\end{tabular} & 1$\times$8 A100 80GB  \\
    Time & – & 72h \\
    \midrule
    \textbf{Cost Est.} & – & \$1088  \\
    
    \midrule
      & \textbf{Qwen2.5-SimpleRL\citep{zeng2025simplerlzooinvestigatingtamingzero}} & \textbf{RPO} \\
     \midrule
    \textbf{Base Model}  & {Qwen2.5-Math-7B} & DS-R1-Distill-1.5B \\
    \midrule
    Hardware & 4$\times$6 A100 80GB & 1$\times$8 A100 80GB \\
    Time & 36h & 7h \\
    \midrule
    \textbf{Cost Est.} & \$1633 & \$56 \\
    \bottomrule
    \end{tabular}
}
\end{table}

\section{More Analysis}

\paragraph{The impact of cache pool update strategies.} To investigate the impact of different cache pool update strategies on model performance, we set $\epsilon$ to 0, 0.1, 0.5, and 1 during training. In addition, we also evaluate the model with cache pool updates completely disabled.
As shown in Table~\ref{tab:epsilon impact}, the performance of the 1.5B model exhibits a trend of first improving and then declining as $\epsilon$ increases. The best performance is achieved when $\epsilon = 0.1$, with an average accuracy of 52.6\%.

\begin{table}[H]
    \centering
    \caption{The impact of $\epsilon$ on model zero-shot performance. In the table, no update denotes the case where the cache pool is not updated, which serves as a baseline for comparison.}
    \label{tab:epsilon impact}
    \renewcommand{\arraystretch}{1.1} 
    \setlength{\tabcolsep}{2pt} 
    \rowcolors{2}{gray!15}{white} 
    \resizebox{\linewidth}{!}{
    \begin{tabular}{l|ccccccc}
    \toprule
    \textbf{$\epsilon$} & \textbf{AIME25} & \textbf{AIME24} & \textbf{MATH500} & \textbf{AMC23} & \textbf{Minerva} & \textbf{OlyB}  & \textbf{Avg} \\
    \midrule
    0   & 23.3 & 30.0 & 84.8 & 75.0 & 28.3 & 52.4 & 50.0 \\
    0.1 & 23.3 & 36.7 & 85.4 & 87.5 & 29.4 & 53.0 & 52.6 \\
    0.5 & 30.0 & 26.7 & 83.8 & 72.5 & 28.3 & 52.4 & 49.0 \\
    0.9 & 26.7 & 36.7 & 84.4 & 70.0 & 29.8 & 53.0 & 50.1 \\
    no update & 26.7 & 30.0 & 82.8 & 75.0 & 29.4 & 51.2 & 49.2 \\
    \bottomrule 
    \end{tabular}
    }
\end{table}

\section{Time Overhead for Cache Pool Initialization}

This section reports whether RPO can still achieve significant training acceleration and performance improvement under extreme conditions, such as when the number of epochs is only 1.

\begin{table}[H]
\centering
\caption{Cache pool initialization time (minutes) for 1.5B and 7B models under different GPU types, dataset sizes, and GPU counts.}
\label{tab:combined_time_comparison}

\centering
\small 
\caption*{1.5B Model}
\resizebox{\linewidth}{!}{
\begin{tabular}{lcccc}
\toprule
\textbf{Dataset} & \textbf{GPU} & \textbf{1} & \textbf{4} & \textbf{8} \\
\midrule
\multirow{2}{*}{7k}   & H20  & 34.78 & 24.61 & 15.10 \\
                       & A100 & 32.11 & 21.45 & 13.98 \\
\midrule
\multirow{2}{*}{70k}  & H20  & 347.91 & 249.14 & 160.87 \\
                       & A100 & 327.19 & 214.78 & 135.89 \\
\bottomrule
\end{tabular}
}

\vspace{4mm}

\caption*{7B Model}
\resizebox{\linewidth}{!}{
\begin{tabular}{lcccc}
\toprule
\textbf{Dataset} & \textbf{GPU} & \textbf{1} & \textbf{4} & \textbf{8} \\
\midrule
\multirow{2}{*}{7k}   & H20  & 58.57 & 26.44 & 17.95 \\
                       & A100 & 55.43 & 22.56 & 16.19 \\
\midrule
\multirow{2}{*}{70k}  & H20  & 582.95 & 261.28 & 179.13 \\
                       & A100 & 566.49 & 238.91 & 167.57 \\
\bottomrule
\end{tabular}
}
\end{table}

Table ~\ref{tab:combined_time_comparison} shows the model initialization time for the 1.5B and 7B models under different GPU count configurations. Table ~\ref{tab:training_time_comparison} shows the training time for one epoch on an 8-card H20 machine and an 8-card A100 machine, including the computational overhead of cache initialization. As seen from the table, even in extreme cases with only a single epoch of training, RPO can still provide significant acceleration.

\begin{table}[H]
    \centering
    \caption{Training time comparison (in hours) of DeepSeek-R1-Qwen models on H20 and A100 GPUs.}
    \label{tab:training_time_comparison}
    \resizebox{\linewidth}{!}{
    \begin{tabular}{l|c|c}
        \toprule
        \textbf{Model} & \textbf{H20 (hours)} & \textbf{A100 (hours)} \\
        \midrule
        \textbf{DS-R1-Qwen-1.5B} & & \\
        \quad RPO & 14.45 & 12.41 \\ 
        \quad GRPO & 40.98 & 37.35 \\
        \midrule
        \textbf{DeepSeek-R1-Qwen-7B} & & \\
        \quad RPO & 39.64 & 37.38 \\ 
        \quad GRPO & 114.50 & 105.70 \\
        \bottomrule
    \end{tabular}
    }
\end{table}

\section{Pseudo Code for RPO Training Process}

The pseudo code of \emph{Replay-based Policy Optimization}(RPO) in the training process is as follows.

\begin{algorithm}
\small 
\setstretch{0.9} 
\caption{Compact RPO Training}
\KwIn{Dataset $\mathcal{D}$, model $\pi_{\theta_0}$, cache $\mathcal{C}^{(0)}$, params $\eta,\beta,\epsilon,G$}
\KwOut{Model $\pi_{\theta_T}$, cache $\mathcal{C}^{(T)}$}

\textbf{Initialize Cache:} \\
$\mathcal{C}^{(0)}=\emptyset$

\For{$q_k \in \mathcal{D}$}{
    $\mathcal{C}^{(0)} \gets \mathcal{C}^{(0)} \cup \{(q_k, \pi_{\theta_0}(\cdot|q_k))\}$ 
}

\For{$t \gets 1$ \KwTo $T$}{
    \textbf{Rollout:} \\
    \For{$q_k \in \mathcal{D}$}{
        $a_k \gets  \{ a \mid (q_k, a) \in \mathcal{C}^{(t-1)} \}$ \\
        \For{$g \gets 1$ \KwTo $G$}{
            $\tilde{a}_i^{(g)} \gets \text{Concat}(a_k^{[0:-m]}, \pi_{\theta_{t-1}}(\cdot|q_k, a_k^{[0:-m]}))$ \\
            $R_i^{(g)} \gets r(q_k, \tilde{a}_i^{(g)})$
        }
    }
    
    \textbf{Optimize:} \\
    $\theta_t \gets \theta_{t-1} + \eta \nabla_{\theta}J_{\mathrm{GRPO-Cache}}$ \\
    
    \textbf{Update Cache:} \\
    \For{$q_k \in \mathcal{D}$}{
        \eIf{$u \leq \epsilon$}{
            $ \mathcal{C}^{(t-1)} \cup \left\{(q_k, o_{\text{argmax} \{R_i\}^{G}_{i=1}})\right\}  \setminus \left\{(q_k, a_k)\right\}$ 
        }{
            $\mathcal{C}^{(t)} \gets \mathcal{C}^{(t-1)} \cup \{(q_k, o_{g'})\} \setminus {(q_k,a_k)}$ \tcp*{$g'\sim\mathcal{U}(1,G)$}
        }
    }
}
\end{algorithm}
\end{document}